\documentclass[10pt,twocolumn,letterpaper]{article}

\usepackage[dvipsnames]{xcolor}
\usepackage[pagenumbers]{cvpr} % To force page numbers, e.g. for an arXiv version

\usepackage[dvipsnames]{xcolor}

\usepackage{times}
\usepackage{epsfig}
\usepackage{graphicx}
\usepackage{amsmath}
\usepackage{amssymb}
\usepackage{bbm}
\usepackage{booktabs}
\usepackage{bm}
\usepackage{multirow}
\usepackage{microtype}
\usepackage[T1]{fontenc}
\usepackage{soul}

\usepackage{xcolor}
\usepackage{listings}
\usepackage{paralist}
\usepackage[toc,page]{appendix}

\definecolor{cvprblue}{rgb}{0.21,0.49,0.74}
\def\MethodName{\textit{$\mathcal{L}$asagna}}
\def\DataName{\textit{$\mathcal{R}$eliT}}
\usepackage[pagebackref,breaklinks,colorlinks,citecolor=cvprblue]{hyperref}

\newif\ifcomments

\commentstrue

\newcommand{\xhdr}[1]{\vspace{3pt}\noindent\textbf{#1}}

\def\paperID{2270} % *** Enter the Paper ID here
\def\confName{CVPR}
\def\confYear{2024}
\usepackage{bm}
\title{\parbox{0.035\textwidth}{\includegraphics[width=\linewidth]{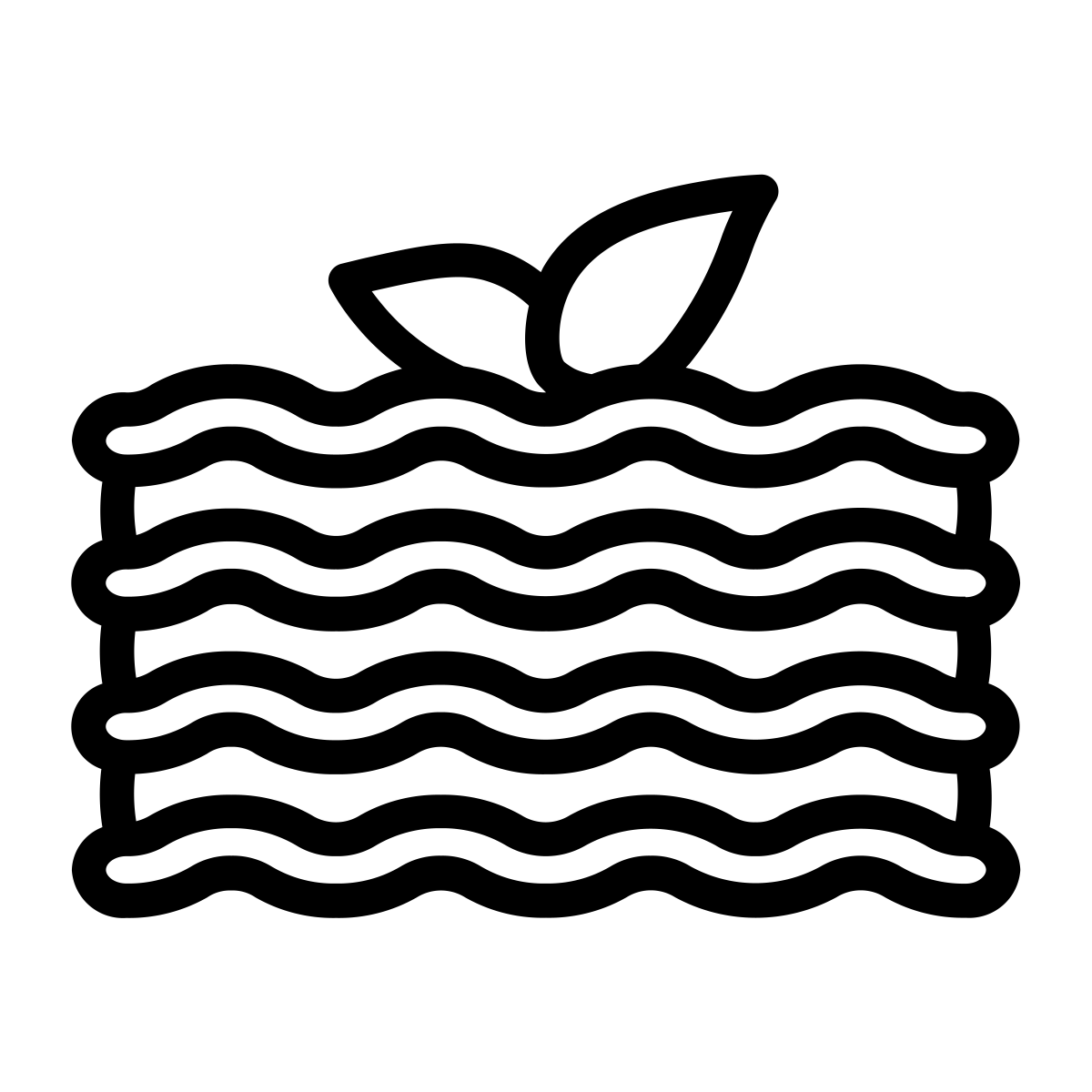}} \MethodName: Layered Score Distillation for Disentangled Object Relighting}  % **** Enter the paper title here
\author{Dina Bashkirova\textsuperscript{1}\\
{\tt\small dbash@bu.edu}
\and
Arijit Ray\textsuperscript{1}\\
{\tt\small array@bu.edu}
\and
Rupayan Mallick\textsuperscript{2} \\
{\tt\small rm2083@georgetown.edu}
\and
Sarah Adel Bargal\textsuperscript{2} \\
{\tt\small sb2122@georgetown.edu }
\and 
Jianming Zhang\textsuperscript{4} \\
{\tt\small jianmzha@adobe.com}
\and
Ranjay Krishna\textsuperscript{3} \\
{\tt\small ranjay@cs.washington.edu}
\and
Kate Saenko\textsuperscript{1,5}\\
{\tt\small saenko@bu.edu}}

\begin{document}
\maketitle

\footnotetext[1]{Boston University}
\footnotetext[2]{Georgetown University}
\footnotetext[3]{University of Washington}
\footnotetext[4]{Adobe Reaearch}
\footnotetext[5]{Facebook AI Research}

\begin{abstract}
    
Professional artists, photographers, and other visual content creators use object relighting to establish their photo's desired effect. 
Unfortunately, manual tools that allow relighting have a steep learning curve and are difficult to master;
Although generative editing methods now enable some forms of image editing, relighting is still beyond today's capabilities; existing methods struggle to keep other aspects of the image—colors, shapes, and textures—consistent after the edit.
We propose \textbf{\MethodName}, a method that enables intuitive text-guided relighting control. 
\MethodName~learns a lighting prior by using score distillation sampling to distill the prior of a diffusion model, which has been finetuned on synthetic relighting data.
To train \MethodName, we curate a new synthetic dataset \textbf{\DataName}, which contains 3D object assets re-lit from multiple light source locations. 
Despite training on synthetic images, quantitative results show that \MethodName~ relights real-world images while preserving other aspects of the input image, outperforming state-of-the-art text-guided image editing methods. 
\MethodName~ enables realistic and controlled results on natural images and digital art pieces and is preferred by humans over other methods in over $91\%$ of cases. 
Finally, we demonstrate the versatility of our learning objective by extending it to allow colorization, another form of image editing. The code for \MethodName~ can be found at \url{https://github.com/dbash/lasagna}. 
\end{abstract}
\section{Introduction}
\begin{figure}[t]
    \centering
    \includegraphics[width=\linewidth]{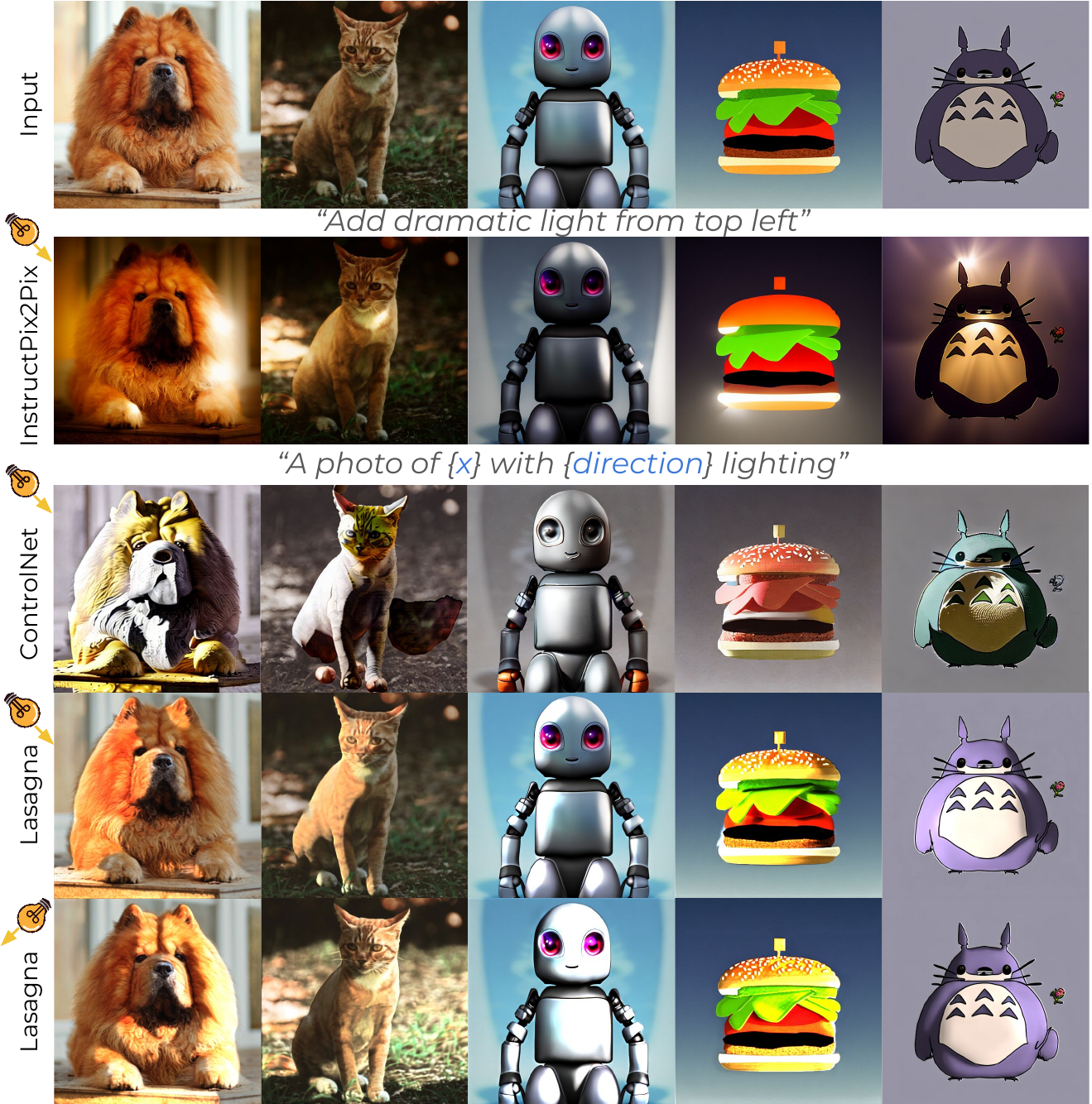}
    \caption{\small{Given a single input image (\textit{top row}) and a text prompt specifying a lighting condition (e.g. light pointing from the left or the right)
    %and right sides in the \textit{bottom rows}, respectively), 
    \MethodName~ performs geometry-aware relighting of images of various levels of realism, from natural photos to minimalist digital art, and allows text-guided light conditioning (\textit{fourth and fifth rows}). State-of-the-art InstructPix2Pix~\cite{brooks2023instructpix2pix} (\textit{second row}), given the prompt that works best in our experiments, struggles to perform realistic relighting, while ControlNet~\cite{zhang2023adding} (\textit{third row}) trained on relighting data often alters crucial aspects of the image.}}
    \label{fig:teaser}
\end{figure}

\begin{figure*}
    \centering
    \includegraphics[width=\linewidth]{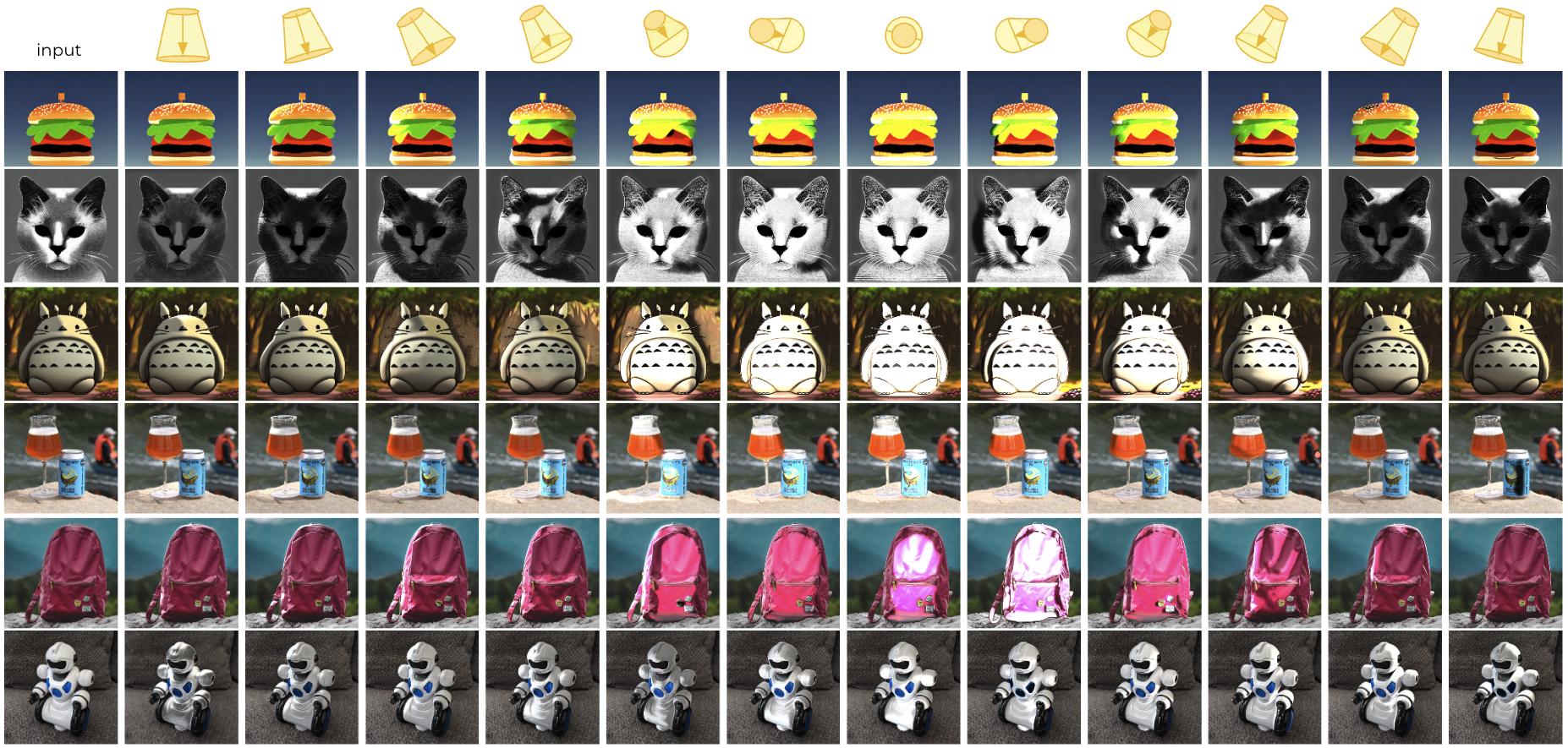}
    \caption{\small With \MethodName, we can perform controlled relighting with language guidance only (light direction shown at the top). Given a simple language prompt that specifies the lighting direction, ~\MethodName~ performs shape-aware relighting and generalizes to both natural images and digital art examples of various levels of abstraction. }
    \label{fig:controlled_res}
\end{figure*}

% Paragraph 1: Setting up the problem.
Professional visual artists spend hours editing their photos. Editing requires a wealth of knowledge about the 3D scene: the geometry of objects, their position, and how the direction of the light source affects their appearance\textbf{~\cite{bomback1971manual,hunter2021light}}. 
Manual editing tools like Photoshop~\cite{photoshop} and Procreate~\cite{procreate} can be tedious and time-consuming for novices and professionals alike. 
While specialized methods for relighting and shading exist~\cite{yeh2022learning, liao2015approximate, zhang2022simbar}, they require supervised training to estimate object intrinsics and therefore lack generalization.
% they rely on complicated pipelines to estimate object intrinsics and lack generalization.

% Paragraph 2: Laying out the technical challenges with model-based editing.
Recent diffusion-based editing methods~\cite{mokady2023null,brooks2023instructpix2pix,kawar2023imagic} promise to offer a more user-friendly editing experience through language interactions and generalize to a variety of image domains. 
Unfortunately, today's methods have limited success in control over lighting, shading or reflection, due to a lack of such geometry-based knowledge in their pre-training.
As shown in Figure~\ref{fig:teaser}, if we instruct a state-of-the-art image editing method InstructPix2Pix~\cite{brooks2023instructpix2pix} to relight images, it fails to control light direction, doesn't apply realistic relighting in general, and also doesn't preserve the overall appearance and content. 
One effective way to improve specific editing capabilities is finetuning with supervision, however, existing datasets for relighting are typically too small to allow generalization~\cite{murmann2019dataset,toschi2023relight}. 
Finetuning on synthetic data has shown to be a promising alternative~\cite{michel2023object} for geometry-aware edits. However, as shown in Figure~\ref{fig:teaser}, directly applying a model, which was trained on synthetic data, for relighting real images leads to undesired changes to other crucial aspects of the input.

% Paragraph 3: Contribution and describe the model's technical contribution.
To provide an intuitive and realistic text-guided relighting solution, we propose \textbf{\MethodName}~ --  a method for subtle geometry-aware image edits guided by a user's language instruction (e.g.~``a photo of a dog with $\{\operatorname{direction}\}$ lighting''). 
There are two key technical contributions behind \MethodName.
First, inspired by DreamFusion~\cite{poole2022dreamfusion}, \MethodName~ learns to extract the geometry prior from a diffusion model using score distillation sampling.
Second, \MethodName~ decomposes relighting as a luminosity adjustment in the pixel space, which allows for a restricted edit that does not alter other aspects of an input image. Together, these two techniques disentangle object relighting from other types of editing.
%and disentangled edit that does not alter other aspects of an input image.
Representing lighting edits as a pixel-wise luminosity adjustment is inspired by tools like Photoshop~\cite{photoshop} and Procreate~\cite{procreate}, thus such layered representation can be especially useful and intuitive for visual content creators.
% With this decomposition, \MethodName{} does not need to explicitly estimate object geometry, and, unlike general text-guided image editing methods, \MethodName{} guarantees preservation of the input image content by restricting editing to the luminosity layers. 

% Paragraph 4: Talk about the dataset contribution.
To train \MethodName, we propose \textbf{\DataName}, a large-scale synthetic dataset of scenes with objects rendered with varying light sources and directions.
\DataName~ is created using the 3D object assets from Objaverse~\cite{deitke2023objaverse} and contains over $14,000$ objects. 
\DataName~training introduces a relighting prior to a diffusion model and to learn a robust association between the language guidance and the lighting direction, disentangling it from other factors of variation. Though this prior is trained using synthetic images only, \MethodName~ leverages it to perform controlled and geometrically accurate relighting on real world images and even digital art pieces.

% Paragraph 5: End with a summarization of the results.
We show examples of \MethodName's outputs in Figure~\ref{fig:controlled_res},
showing it efficacy on datasets from three different domains: natural photos from the Dreambooth dataset, digital art pieces, and the test set of the proposed \DataName~ dataset. Our experimental results indicate a clear advantage of \MethodName~ over state-of-the-art text-guided image editing methods as well as the baseline finetuned for relighting, with our method being preferred in over $91\%$ of cases according to a human evaluation study.
We also illustrate the versatility of \MethodName~ by extending it for sketch colorization.

\section{Related Work}
We situate our work within the space of text-guided image editing. We draw on score distillation sampling to train our model for object relighting.

\vspace{-10pt}\paragraph{Text-guided image editing and translation} became a topic of active research shortly after the emergence of LLM- and VLM-based text-conditional image generation methods, such as Stable Diffusion~\cite{rombach2022high}, Imagen~\cite{saharia2022photorealistic} and DALLE~\cite{ramesh2021zero,ramesh2022hierarchical}. Earlier text-guided editing methods used CLIP~\cite{radford2021learning} to train a separate generator to learn a realistic edit for a given input image~\cite{kwon2022clipstyler,bar2022text2live,gal2022stylegan}, while other methods like SDEdit~\cite{meng2021sdedit}, Plug-and-Play~\cite{Tumanyan_2023_CVPR}, DDIB~\cite{su2022dual} and Imagic~\cite{kawar2023imagic} directly use a diffusion prior to edit the input image.  Prompt2Prompt (P2P)~\cite{hertz2022prompt} performs editing by swapping cross-attention maps, while InstructPix2Pix~\cite{brooks2023instructpix2pix} further uses P2P to generate extract pseudo-supervision for image editing based on a single editing prompt.   
Another direction in image editing is text-guided inpainting~\cite{wang2023imagen,chang2023muse} that edits a masked part of an image according to the guiding prompt. 

While these methods achieve remarkable editing results, the majority of them predict an edited image directly, which often results in undesired alterations in the aspects of the input image that are not supposed to be edited, which is especially critical for geometry-aware edits, such as relighting. A concurrent work, Delta Denoising Score~\cite{hertz2023delta}, uses a version of score distillation sampling to perform few-shot image editing, directly editing the image pixels, however, it underperforms on geometry-aware edits like relighting due to a lack of such knowledge in the diffusion model. Text2Live~\cite{bar2022text2live}  learns a RGB-A edit layer using a CLIP \cite{radford2021learning} prior, which is also limited for geometry-aware edits. In contrast, \MethodName~ uses a relighting fine-tuned diffusion model prior, which leads to more realistic edits, and restricts the editing effect via functional layer composition. To the best of our knowledge, \MethodName~ is the first method that uses a diffusion model prior to perform object relighting without any real training data.

\vspace{-0pt}\paragraph{Score distillation sampling.} (SDS) was introduced in DreamFusion~\cite{poole2022dreamfusion} and then further improved in the follow-up works~\cite{wang2023prolificdreamer,wang2023score,chen2023fantasia3d} for text-conditional 3D object generation with NeRFs~\cite{mildenhall2021nerf}, allowing distillation of a diffusion model prior via differentiable image parameterization. One of the biggest advantages of differential image parameterization is that it allows introduction of custom sampling constraints. Works that use score distillation sampling include a few 2D image generation results~\cite{poole2022dreamfusion,wang2023prolificdreamer,wang2023score} to show text-conditional image generation capabilities, but to the best of our knowledge, we are the first to use \emph{image-conditional} SDS for disentangled image editing.  

\vspace{-0pt}\paragraph{Object relighting} can be roughly divided into inverse rendering-based approaches~\cite{barron2012color,karsch2011rendering,yeh2022learning, liao2015approximate, zhang2022simbar} that decompose the geometric properties of the object, such as reflectance, normals and material, and image editing methods~\cite{xu2018deep,wang2020people,carlson2019shadow,dipta2021msr} that predict the relighting edit directly in the pixel space. While the latter methods often produce inconsistent shading results due to the lack of geometric prior, inverse rendering-based methods are trained on curated datasets and, therefore, struggle to generalize. Recent works on relighting perform 3D scene reconstruction with NeRFs~\cite{boss2021nerd,srinivasan2021nerv,xu2023renerf}, which requires a set of images from different view points and a known light source, limiting it's applicability to a wide variety of domains such as digital art.
In contrast, \MethodName~ requires a single image and leverages the rich prior of a diffusion model, which leads to better generalization to a wide range of image domains and allows a more intuitive way to condition relighting with text prompts.
% Inspired by the success of Object3DIT~\cite{michel2023object} on scene recomposition, we surpass the inverse rendering step and instead perform relighting in the 2D space by leveraging relighting prior learned from a synthetic 3D scene manipulation dataset.

\vspace{-10pt}\paragraph{Relighting datasets.} While there exist some real-world~\cite{murmann2019dataset,toschi2023relight} and synthetic~\cite{helou2020vidit,xu2018deep} relighting datasets, they are relatively small, containing less than $1000$ distinct objects which lead to overfitting. In contrast, our proposed \DataName~ dataset contains more than $13K$ distinct 3D objects from Objaverse~\cite{deitke2023objaverse} and $50$ high-quality background environment maps, drastically increasing diversity compared to the existing datasets. 
\section{Lasagna}
\begin{figure*}
    \centering
    \includegraphics[width=\linewidth]{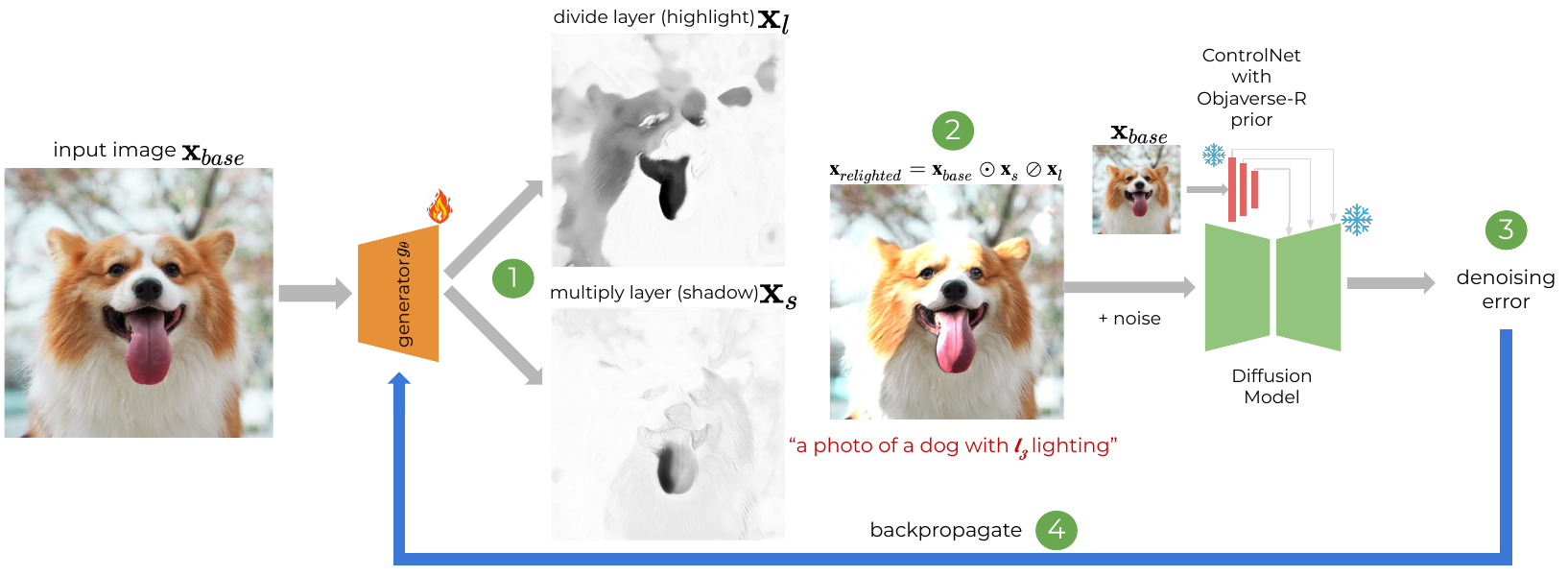}
    \caption{\small Schematic overview of \MethodName~ for relighting. Given an input base layer $\bm{x}_{\operatorname{base}}$, e.g. real photo, \MethodName~ learns an editing layer using score distillation sampling: \textbf{1)} a generator $g_{\theta}$ predicts two relighting layers -- a multiply layer $\bm{x}_{s}$ and a divide layer $\bm{x}_{l}$ that decrease and increase luminosity respectively; \textbf{2)} an input layer $\bm{x}_{\operatorname{base}}$ is composed with the relighting layers $\bm{x}_{l}$ and $\bm{x}_{s}$ into an edited image $\bm{y}$; \textbf{3)} $\bm{y}$ is perturbed with a random noise and passed to a frozen diffusion model with a ControlNet adaptor trained on \DataName~ dataset to compute a denoising error conditioned on the guiding prompt ({\color{BrickRed}in red}) that specifies lighting direction; \textbf{4)} the weighted denoising error is backpropagated to update the generator parameters $\theta$ according to Eq.~\ref{eq:sds}.}
    \label{fig:method}
\end{figure*}

Our goal is to achieve disentangled object relighting with text guidance using a strong diffusion model prior. To achieve this, we first introduce score distillation sampling in Section~\ref{sec:prelim} that allows image generation with custom constraints, we then discuss how score distillation sampling can be used for disentangled image editing by introducing editing layers in Section~\ref{sec:layered}, and explain how to extend the layered editing framework to disentangled relighting. Finally, since off-the-shelf diffusion models lack a fine-grained language prior for lighting, we introduce it via training an image-conditional adaptor on a novel synthetic dataset for controlled relighting, which we discuss in Section~\ref{sec:lighting_prior}.

\subsection{Preliminaries: diffusion and score distillation sampling}
\label{sec:prelim}
Given an input image $\bm{x}\in \mathcal{X}$, and the embeddings  $\tau(c)$ of the conditioning text prompt $c$, a diffusion model learns to predict the noise $\bm{\epsilon}$ added to an image $\bm{x}$  with a denoising model $\hat{\epsilon}_{\phi}$ at a random time step $t$ via weighted denoising score matching: 
\begin{equation}
\label{eq:ldm}
\mathcal{L}_{\operatorname{Diff}} := \mathbb{E}_{\bm{x}\sim \mathcal{X},\bm{\epsilon}\sim\mathcal{N}(0, 1)}[w(t)||\hat{\epsilon}_{\phi}(\bm{x}_t, t, \tau(c)) - \bm{\epsilon} ||^2_2] \end{equation}
 where $\bm{x}_t = \alpha_t\bm{x} + \sigma_t\bm{\epsilon}$ and $w(t)$ is a weighting function dependent on the timestep $t$.
 DreamFusion~\cite{poole2022dreamfusion} introduced score distillation sampling (SDS) to train a NeRF~\cite{mildenhall2021nerf} representing a 3D object using the 2D prior of a diffusion model. To learn a 3D representation, DreamFusion uses differential parametrization~\cite{mordvintsev2018differentiable} -- a general framework in which parameters $\theta$ of a differential generator $g$ are trained to produce an image $\bm{x} = g(\theta)$ minimizing a loss function $f(\bm{x})$, where $f$ is typically computed based on the output or features of another model. This formulation enables the use of an arbitrary generator function $g$ -- in case of DreamFusion, a NeRF -- and allows to incorporate additional generation constraints. To train a differentiable generator $g$, DreamFusion proposes to use the following gradient update rule:
 \begin{equation}
    \nabla_{\theta} \mathcal{L}_{\operatorname{SDS}} = \mathbf{E}_{t, \bm{\epsilon}} [w(t)(\hat{\epsilon}_{\phi}(\bm{x}_t, t, \tau(c)) - \bm{\epsilon})\frac{\partial \bm{x}}{\partial\theta}]
    \label{eq:sds}
 \end{equation}
where in case of DreamFusion, $g$ is a differential renderer, and $\theta$ are the parameters of an MLP representing a NeRF.

\begin{figure*}
    \centering
    \includegraphics[width=1.\textwidth]{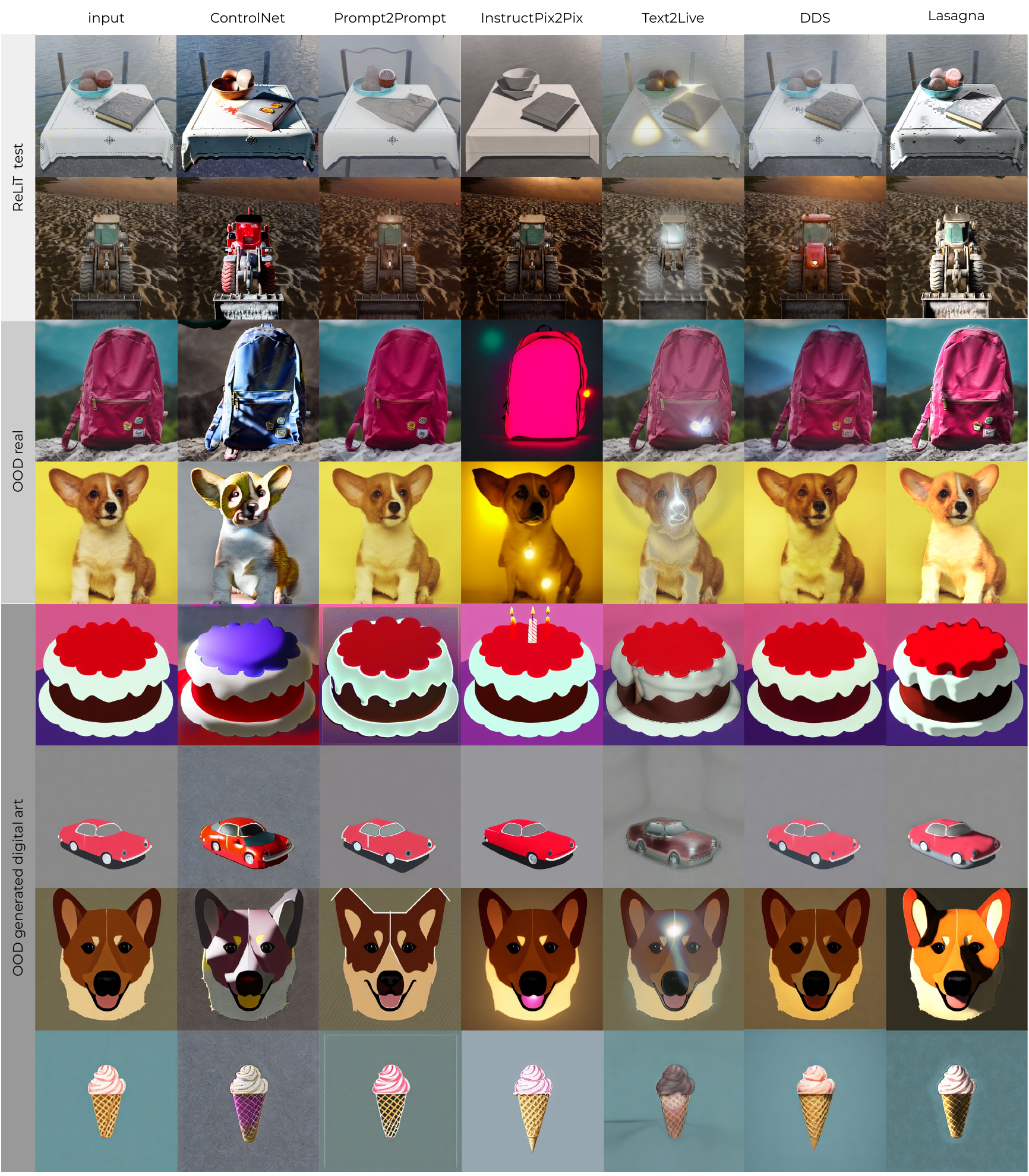}
   \caption{\small Comparison of shading results of the baseline methods: ControlNet trained on \DataName~, Prompt2Prompt~\cite{hertz2022prompt}, InstructPix2Pix~\cite{brooks2023instructpix2pix}, Text2Live~\cite{bar2022text2live}, DDS~\cite{hertz2023delta} and \MethodName~ on the test set of \DataName~, out-of-distribution (OOD) real photos from the DreamBooth dataset~\cite{ruiz2023dreambooth} and digital art examples generated by Stable Diffusion v2.1. More results can be found in Section~\ref{sec:app_baselines} in Appendix. }
    \label{fig:shading_main}
\end{figure*}
\subsection{Layered score distillation sampling}
\label{sec:layered}
Given an input image $\bm{x} \sim \mathcal{X}$ and an editing text prompt $c$, our goal is to predict an edited version of $\bm{x}$, $\bm{y}$, s.t. $\bm{y}$ follows the editing prompt $c$ while preserving all other aspects of $\bm{x}$ fixed. Since directly predicting the editing result $\bm{y}$ from $\bm{x}$ often leads to changing the aspects of $\bm{x}$ unrelated to the edit (as shown in Figures~\ref{fig:objaverse_train_ex} and~\ref{fig:shading_main}), we propose a \emph{layered} approach in which a separate image -- referred to as the \textit{editing layer} in the remainder of the paper -- is predicted and composed with the input image via a fixed edit-specific layer composition function $f: \mathcal{R}^{H\times W\times C_1}, \mathcal{R}^{H\times W\times C_2} \rightarrow \mathcal{R}^{H\times W\times C_3}$ to achieve editing results.  An input base layer  $\bm{x}_{\operatorname{base}}$ is passed to a generator $g_{\theta}$ parametrized by $\theta$ (in our experiments -- UNet~\cite{ronneberger2015u}, analogously to Text2Live~\cite{bar2022text2live}) that predicts an editing layer $\bm{x}_{\operatorname{edit}}$ (this approach trivially extends to multiple editing layers, we focus on a single layer for simplicity). Then, $\bm{x}_{\operatorname{base}}$ and  $\bm{x}_{\operatorname{edit}}$ are composed into an editing result $\bm{y} = f(\bm{x}_{\operatorname{base}}, \bm{x}_{\operatorname{edit}})$ that is passed to a frozen diffusion model to compute a denoising error from Eq.~\ref{eq:sds} for a randomly sampled time step $t$. Finally, the denoising error is backpropagated to update the parameters $\theta$. An additional regularization function can be added to the denoising loss to enforce additional properties or constraints of an edit.

\paragraph{Layered editing for shading and relighting.}
Inspired by the relighting functionality proposed in image editing tools like Procreate, we define relighting as a luminosity adjustment function. The generator $g_{\theta}$ predicts two editing layers -- a shading and a lighting layer $\bm{x}_{s}, \bm{x}_{l} \sim \mathcal{R}^{H \times W \times 1}$, with $0 \leq \bm{x}_{s}, \bm{x}_{l} \leq 1$ -- that adjust the input image luminosity via Hadamand product and division, respectively. The underlying composition function is defined as $f_{\operatorname{relighting}}(\bm{x}_{\operatorname{base}}, \bm{x}_s, \bm{x}_l) =  \bm{x}_{base} \odot \bm{x}_s \oslash \bm{x}_l$. To introduce a minimal change in luminosity, we add a regularization loss $L_{\operatorname{reg}}(\bm{x}) = ||1 - \bm{x}||_1$ for both layers. An overview of the \MethodName~ relighting is illustrated in Figure~\ref{fig:method}.
The final gradient update rule is as follows:
 \begin{equation}
    \nabla_{\theta} \mathcal{L}_{\operatorname{SDS}} = \mathbf{E}_{t, \bm{\epsilon}} [(w(t)(\hat{\epsilon}_{\phi}(\bm{x}_t, t, \tau(c)) - \bm{\epsilon}) + L_{\operatorname{reg}}(\bm{x})]\frac{\partial \bm{x}}{\partial\theta}
 \end{equation}
The resulting editing layers can \emph{only adjust luminosity} by design, preserving other aspects of an input image unchanged.
\begin{figure*}
    \centering
    \includegraphics[width=\linewidth]{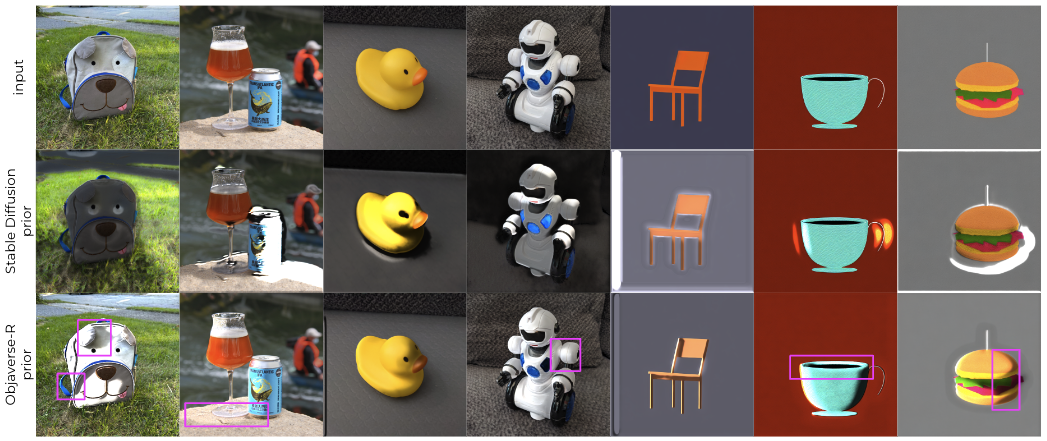}
    \caption{\small Ablation study results on layered score distillation sampling with \DataName~ prior compared to StableDiffusion v2.1 prior. Prompt-based relighting guidance with a pretrained Stable Diffusion leads to unreliable relighting results and introduces artifacts. Finetuning a ControlNet on \DataName~ introduces a strong association between the guidance text prompt and the lighting, which results in a substantially more reliable relighting edit. }
    \label{fig:ablation_prior}
\end{figure*}

\subsection{Introducing language-guided lighting prior to diffusion models}
\label{sec:lighting_prior}
As shown in Figure~\ref{fig:ablation_prior}, lighting is an aspect of an image that is hard to control via language with an off-the-shelf pretrained diffusion model due to the limitations in the language prior. Therefore, we introduce additional lighting prior by finetuning a ControlNet~\cite{zhang2023adding} adaptor on a novel \DataName~ dataset based on the popular synthetic Objaverse~\cite{deitke2023objaverse} dataset of 3D models. The ControlNet~\cite{zhang2023adding} adaptor enables image and text conditioning, which allows relight an input image conditioned on a text prompt specifying lighting parameters. 

\noindent\paragraph{\DataName~ dataset} is generated using a subset of ``thing" objects from the Objaverse dataset with a few notable modifications in the rendering pipeline: 1) we applied randomly sampled realistic background maps from $50$ HDRI images collected from \url{https://polyhaven.com/} distributed with a CC0 license; 2) for each object in the dataset, we kept the camera position fixed; 3) we rendered frames with spot light source placed in one of the $12$ locations around the object. Each of the $12$ locations is fixed for the entire dataset, please see Figure~\ref{fig:objaverser} in the Appendix for an illustration of the examples from \DataName~. In addition to $12$ relighting images per object, we collected renderings with the panel light source that achieves uniform lighting with minimal shadows.
We collected $13975$ training object examples and $164$ testing examples. \DataName~ will be made publicly available upon acceptance of this manuscript.

\begin{figure}[t]
    \centering
    \includegraphics[width=\linewidth]{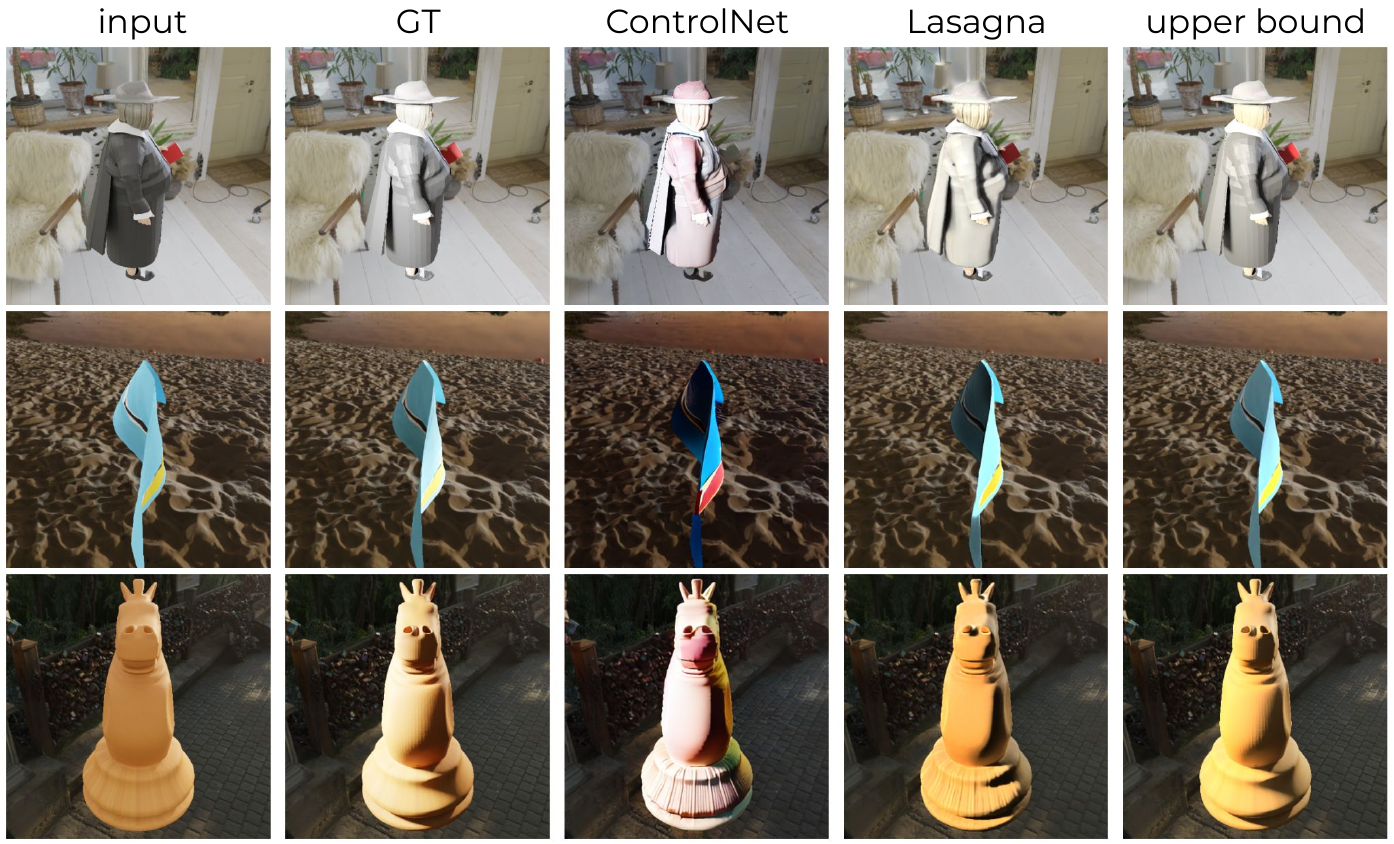}
    \caption{Relighting results on the training set examples of \DataName~ with ControlNet indicate that the model learned to relight the input object given the input light source location index comparably well but struggles to preserve other aspects of the input image, such as colors and shape, fixed. In contrast, \MethodName~ that distills the lighting prior from the same ControlNet adaptor achieves relighting on par with ControlNet while keeping other aspects of the input image.}\vspace{-10pt}
    \label{fig:objaverse_train_ex}
\end{figure}

\vspace{-10pt}\paragraph{Training ControlNet with \DataName.} To introduce a robust language guidance for lighting, we trained a ControlNet adaptor~\cite{zhang2023adding} that allows image conditioning for diffusion models. Given an input image $\bm{x}_{u}$ from \DataName~ with uniform lighting, a ground truth image $\bm{x}_i$ of a corresponding object with the spot light at location $p_i$, and a guidance text prompt $c = \text{``A photo of a \{category\} with \{i\} lighting"}$, with the lighting position index $i$, the denoising model with a ControlNet adaptor $\hat{\bm{\epsilon}}_{\operatorname{c}}$ parametrized by $\theta_c$ is trained via a weighted score matching loss with image conditioning: 
\begin{equation}
    \mathcal{L}_{\operatorname{ControlNet}} := \mathbb{E}_{\bm{x}_{u},\bm{\epsilon}}[w(t)|| \bm{\epsilon} - \hat{\bm{\epsilon}}_{\operatorname{c}}(\bm{x}_{i, t} | t, \tau(c), \bm{x}_{u})||^2_2 ]
\label{eq:ldm_controlnet}
\end{equation}
where $\bm{x}_{i, t} = \alpha_t\bm{x}_t + \sigma_t\bm{\epsilon}$.
By minimizing the denoising loss in Eq.~\ref{eq:ldm_controlnet}, ControlNet learns to perform relighting of an input image $\bm{x}_{u}$ corresponding to the light source location specified in the conditioning text prompt. However, as shown in ~Figure~\ref{fig:objaverse_train_ex}, ControlNet fails to preserve other aspects of the image, e.g. changes the colors or the shape of the input object and background, therefore, we propose  \MethodName~ for disentangling the relighting prior from ControlNet.

\section{Experiments}
\label{sec:results}

% \ranjay{One thing that I find really useful to do when writing the experiments section is the following: start the section with a overall description of what kinds of experiments you will conduct and lay out the main takeaways for the reader.
% %
% Next, make sure that in your results section, each of the main takeaway sentences are highlighted in bold and are the first sentence of the paragraph or even headers to subsections. Check out this paper as an example: \href{https://openaccess.thecvf.com/content/CVPR2023/papers/Ma_CREPE_Can_Vision-Language_Foundation_Models_Reason_Compositionally_CVPR_2023_paper.pdf}{paper link}
% }

\subsection{Setup}
\label{sec:setup}

We use Stable Diffusion v1.5~\cite{rombach2022high} in all our relighting experiments, and we augment it with a ControlNet~\cite{zhang2023adding} adaptor to finetune on \DataName~. We finetune the model for $250,000$ iterations on a single $A6000$ GPU with batch size $4$, which takes about $36$ hours. We use the input prompt \textit{``A photo of \{category\} with $l$ lighting"}, with the category name provided in the Objaverse metadata and $l$ being a light source position index in $0-11$, going counter-clockwise around an object, $0$ being placed directly atop the object, $3$ being placed top left w.r.t the object, $6$ facing the object, $9$ being placed top right, the light source pointed at the object in all setups. During training, A ControlNet is given a conditioning image of the object with a panel light and a guiding prompt and is trained to denoise the corresponding ground truth relighted image. 

For relighting layered SDS with \MethodName, we use the open-source implementation~\footnote{https://github.com/ashawkey/stable-dreamfusion} of DreamFusion to train a UNet~\cite{ronneberger2015u} with a sigmoid activation function to produce two single-channel layers -- a multiply layer for shading and a divide layer for lighting. We use the AdamW optimizer with learning rate $5 \times 10^{-3}$ and a classifier-free guidance scale of $10$, regularization weight $1$, and train for $700$ iterations on a single input image, which takes about $140$ seconds. We set the range of timesteps used for score distillation sampling to $0.02 - 0.98$. 
Additionally, we provide proof-of-concept sketch-to-digital art translation results with alpha layer composition in Section~\ref{sec:results}.

\vspace{-15pt}\begin{figure*}[ht]
    \centering

    % Subfigure 1
    \begin{subfigure}{0.33\textwidth}
    \centering
        \includegraphics[width=\textwidth]{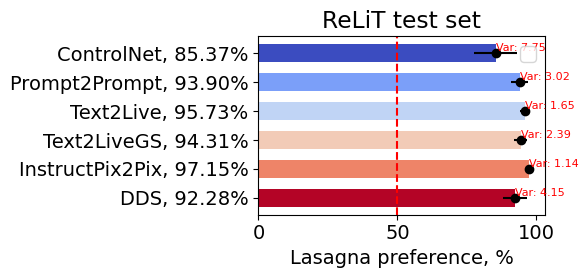}
        % \caption{\DataName~ test set}
    \end{subfigure}
    % Subfigure 2
    \begin{subfigure}{0.33\textwidth}
        \centering
        \includegraphics[width=\textwidth]{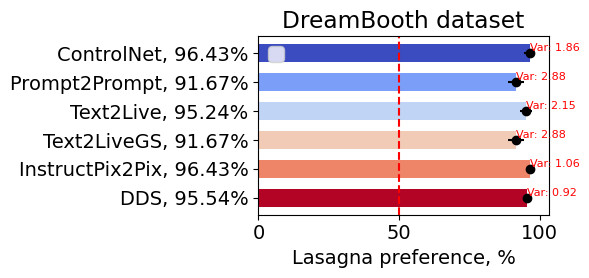}
        % \caption{DreamBooth dataset}
    \end{subfigure}
    % Subfigure 3
    \begin{subfigure}{0.33\textwidth}
        \centering
        \includegraphics[width=\textwidth]{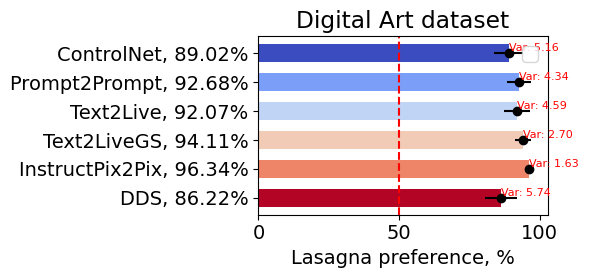}
        % \caption{Digital Art}
    \end{subfigure}

    \caption{\small Percentage of cases in which the participants chose \MethodName~ over each baseline considered in the human study. More details and illustrations of the human evaluation results can be found in Section~\ref{sec:app_human_eval} in the Appendix. }
    \label{fig:human_study}
\end{figure*}

\vspace{20pt}\subsection{Datasets}
\label{sec:datasets}
We finetune a ControlNet adaptor based on a Stable Diffusion model on our novel \DataName~ dataset of $13,975$ objects dataset as described in Section~\ref{sec:lighting_prior}. We test the precise relighting quality of \MethodName~ and compare it with the baseline text-guided editing methods on $164$ test examples from \DataName~ (see Section~\ref{sec:app_objaverser} in Appendix for more details).

To test the relighting capability on natural images, we use the examples from the DreamBooth~\cite{ruiz2023dreambooth} dataset that has photos of $28$ different objects on various backgrounds. We select one image per object, choosing the examples with the most uniform light. To analyze generalization to other domains, we test our method on $164$ digital art images of $13$ categories generated with Stable Diffusion v2.1. with the input prompt \textit{``A minimal digital art of a \{category\}''}, which we refer to as the Digital Art dataset in this paper (more details in Section~\ref{sec:app_digital} in Appendix).

\subsection{Baselines}
\label{sec:baselines}
We compare the quality of relighting of \MethodName~ with the popular text-guided image editing methods: ControlNet~\cite{zhang2023adding} trained on \DataName~, Prompt2Prompt (P2P)~\cite{hertz2022prompt}, Text2Live~\cite{bar2022text2live}, InstructPix2Pix~\cite{brooks2023instructpix2pix} and DDS~\cite{hertz2023delta}. 
% ControlNet is an adaptor-based image-to-image translation method that allows image conditioning for diffusion model inference. Given an image with uniform lighting as conditioning input and a prompt ``A photo of \{category\} in \{X\} lighting", where \textit{category} is the input object category and $X$ is the light source position index, the ControlNet adaptor is trained to relight an input image via the denoising loss in Eq.~\ref{eq:ldm_controlnet}.  
The closest baseline approaches to \MethodName~ are Text2Live that distills CLIP prior into an editing layer and DDS that uses a variant of SDS to edit an image inplace. Text2Live uses an additional green-screen loss to improve the edit fidelity, which is not meaningful for edits such as relighting, therefore, for a fair comparison, we report the Text2Live results both with and without the use of the green-screen loss (Text2LiveGS and Text2Live).  We use the default image editing parameters provided in the official implementation. For all the baselines, we select the prompts and hyperparameters that yield the best relighting results via visual inspection (see details in Section~\ref{sec:app_baselines} of Appendix).

\vspace{-0mm}
\subsection{Evaluation metrics}
\label{sec:metrics}
% For the experiments on the \DataName~ dataset with ControlNet and \MethodName~ where the ground truth relighting data is available, we measure relighting accuracy directly using the Mean Squared Error (MSE) between the ground truth relighted image and the edit result, as illustrated in Figure~\ref{fig:objaverse_train_ex} and Table~\ref{tab:objaverse}. These results show the accuracy of relighting given a precise light source location. 
Our experimental results on the test set of \DataName~ with standard distance metrics such as MSE (see Table~\ref{tab:objaverse} in the Appendix) indicate that pixel-wise metrics do not accurately reflect the relighting quality, as they favor the methods that return an image almost identical to the input and are not sensitive to the relighting quality. Standard relighting metrics such as a re-rendeding error~\cite{liao2015approximate} are not suitable for our setup since they are based on the estimated object intrinsics, therefore, we report the human evaluation results for all datasets using a Two-alternative Forced Choice (2AFC) protocol. Given an input image, a pair of relighting results produced by \MethodName~ and a randomly chosen baseline, the participants were asked  \textit{``Given an input image, which of the following images presents a more realistic relighting of the input object, with light coming from top left?"}. We collected $3$ evaluations per random pair from each of the datasets discussed in Section~\ref{sec:datasets}, a total of $5684$ evaluations, and present the percentage of votes in favor of \MethodName~ compared to each baseline in Figure~\ref{fig:human_study} (more details about human evaluation can be found in Section~\ref{sec:app_human_eval} of the Appendix).

\subsection{Results}
\label{sec:results}
The human evaluation results reported in Figure~\ref{fig:human_study} indicate a clear advantage of \MethodName~ over the baselines that use an off-the-shelf diffusion model and a ControlNet pipeline trained for relighting. The participants prefer \MethodName~ over ControlNet in more than $91\%$ of the cases on the DreamBooth dataset, more than $86\%$ on the Digital Art dataset, and in more than $85\%$ of the cases on the test set of \DataName~ on which ControlNet was trained on, which indicates the importance of prior distillation. \MethodName~ also outperforms the closest baselines -- Text2Live for layered editing and DDS that uses score distillation sampling -- with a $92\%$ preference on \DataName~, $86\%$ preference on Digital Art and $91\%$ on the DreamBooth examples, suggesting that all of the components introduced in \MethodName~ are crucial for high-quality relighting. \MethodName~ is also preferred to state-of-the-art P2P and InstructPix2Pix in more than $91\%$ cases, suggesting the efficiency compared to general image editing methods.
The qualitative examples in Figure~\ref{fig:shading_main} show that \MethodName~ performs relighting on-par with the finetuned ControlNet and superior to the text-guided image editing methods that use an off-the-shelf diffusion model or CLIP, which indicates that adding a text-grounded relighting prior is crucial for achieving controlled relighting. Unlike ControlNet and InstructPix2Pix that alter input image colors, textures and shapes, both Text2Live and \MethodName~ perform editing with separate editing layers, and therefore they both preserve all other aspects of the input image. 

Notably, the relighting prior from a synthetic \DataName~ dataset generalizes well to the images from other domains, such as natural images from DreamBooth and minimal digital art examples from the Digital Art dataset, which can be beneficial for various creative applications. Another important difference of \MethodName~ that makes it stand out compared to the baselines is that it does not require tedious prompt tuning thanks to \DataName~ finetuning on a single prompt template, and it has a relatively small number of hyperparameters (CFG and regularization scale). 
\begin{figure}[t]
    \centering
    \includegraphics[width=\linewidth]{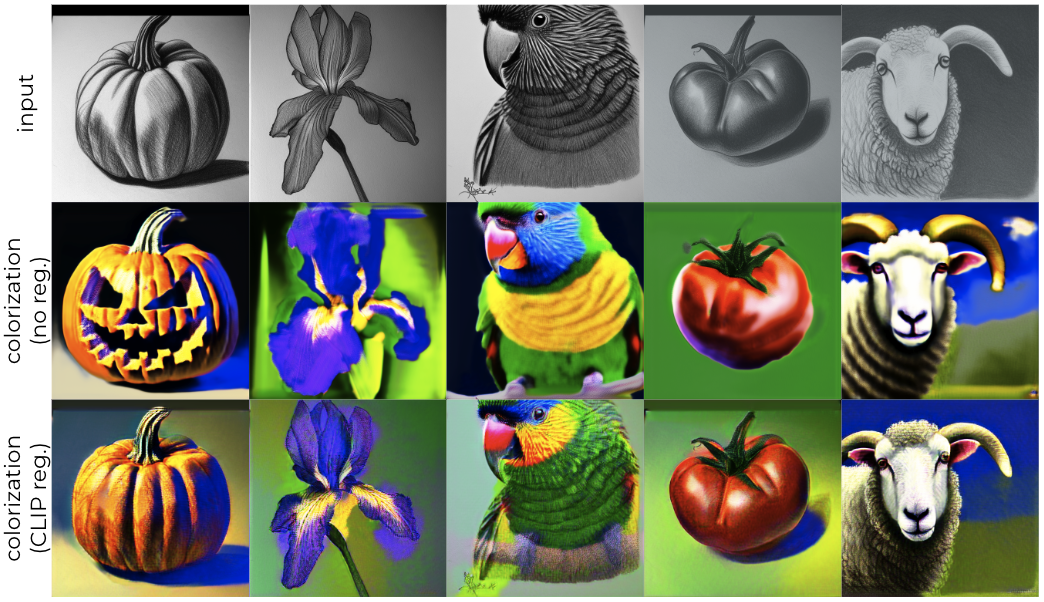}
    \vspace{-15pt}\caption{Alpha layer colorization results for sketch-to-digital art translation with the guiding prompt \textit{``A realistic digital art of an \{X\}"}, where $X$ is the object class name. } \vspace{-10pt}
    \label{fig:colorization}
\end{figure}
\begin{figure}[b]
    \centering
    \includegraphics[width=\linewidth]{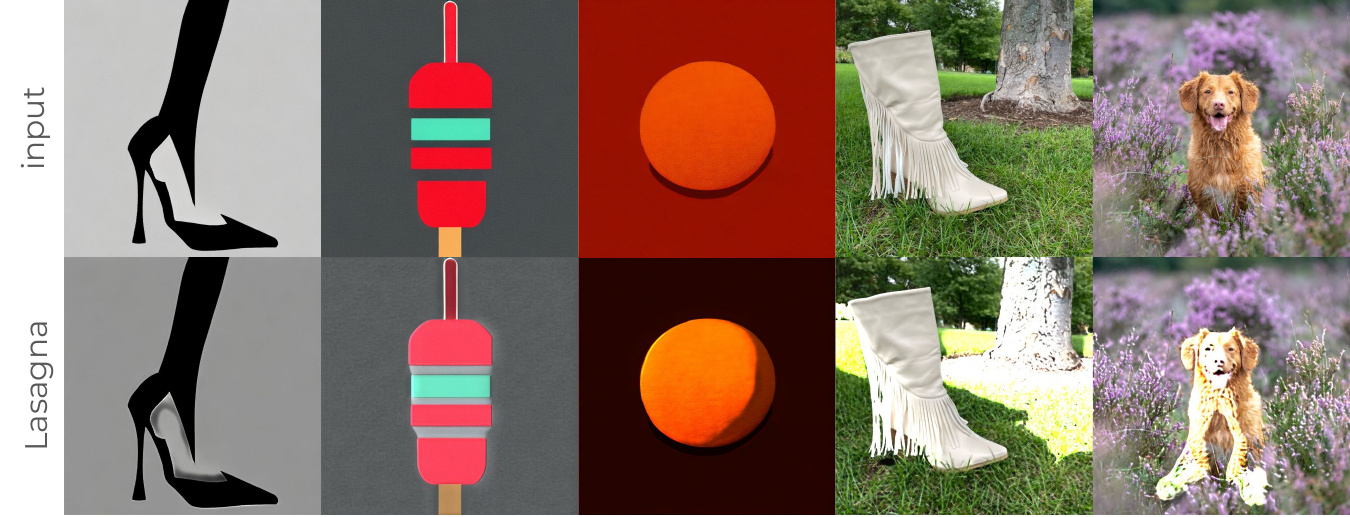}
    \caption{\small Failure cases of \MethodName.}
    \label{fig:lim}
\end{figure}

\vspace{-5pt}\paragraph{Layered editing for colorization.}
To illustrate the versatility of the proposed prior distillation approach for image editing, we show some qualitative results on sketch colorization in Figure~\ref{fig:colorization}. 
We use the prior of an off-the-shelf diffusion model and modify our pipeline as follows: 1) generator $g_{\theta}$ generates two editing layers -- a colorization layer $\bm{x}_{\operatorname{rgb}}$ and a transparency layer $\bm{x}_{\operatorname{\alpha}}$; 2) the layer composition function $f_{\operatorname{color}}$ is an elementwise weighted combination of the input layer and the colorization layer: $f_{\operatorname{color}} = \bm{x}_{\operatorname{\alpha}} \circ \bm{x}_{\operatorname{rgb}} + (1 - \bm{x}_{\operatorname{\alpha}}) \circ \bm{x}_{\operatorname{base}}$. To improve edit quality and diversity, we swap SDS for variational score distillation (VSD)~\cite{wang2023prolificdreamer}; we use a CLIP image feature similarity~\cite{vinker2022clipasso,bashkirova2023masksketch} for structure reqularization (more details on this in Section~\ref{sec:app_colorization} in the Appendix).
These results indicate that we can achieve spatial conditioning with score distillation sampling and the proposed layer composition approach, which can be further improved via regularization. 

\paragraph{Limitations}
As shown in Figure~\ref{fig:lim}, \MethodName~ sometimes fails on the very abstract input images, and is prone to adding some over-exposure lighting artifacts in the background, a limitation that can be mitigated via foreground masking, which is a potential direction for future work.

\vspace{-1mm}
\section{Conclusion}
We show that other text-guided image editing methods fail at relighting due to the limitations in the diffusion model prior, and finetuning on a synthetic relighting dataset helps introduce the prior necessary for relighting, yet it leads to a sim-to-real domain gap altering other aspects of the image. We present a layered image editing method, \MethodName, that disentangles a relighting prior from a model fine-tuned on synthetic data to mitigate the sim-to-real gap and allow a high-quality relighting of images of arbitrary domains. 

% \MethodName~ is a general method for a disentangled image editing that relies on layered score distillation sampling.

\newpage

{
    \small
    \bibliographystyle{ieeenat_fullname}
    \bibliography{main}
}
\newpage
\clearpage

\begin{figure*}
    \centering
    \includegraphics[width=\textwidth]{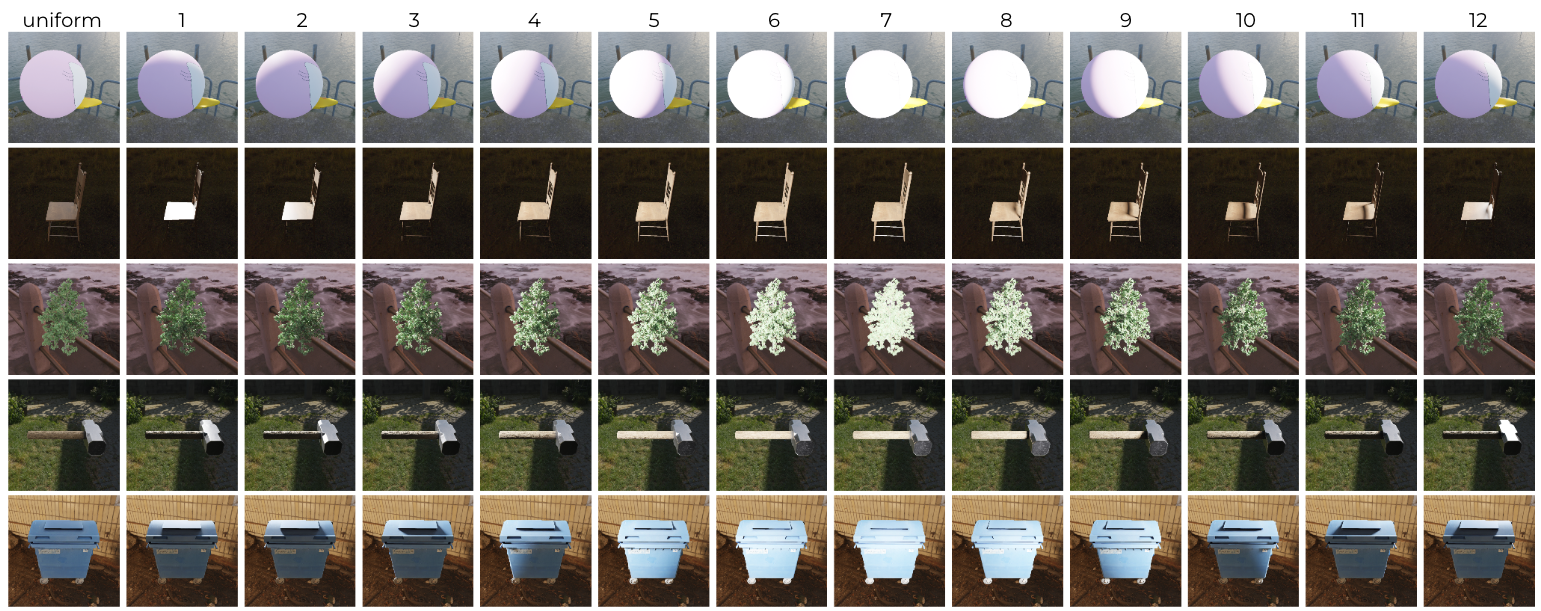}
    \caption{Examples from the \DataName~ dataset: uniformly lit image in the first column, and the same object with spot light source placed at each of the $12$ prederined locations around it.}
    \label{fig:objaverser}
\end{figure*}

\section{Implementation and method details}
\label{sec:app_method_details}
\subsection{Relighting}
\label{sec:app_relighting}
For relighting, we use a version of a convolutional UNet model with a single encoder and two decoder branches, one that produces a highlight layer, and another one predicts a shading layer. The encoder consists of $4$ convolutional layer blocks with $16$, $32$, $64$ and $128$ filters, respectively. Both branches use a final sigmoid activation function and the final values are clipped to the range $[0.1, 1]$ to avoid numerical overflow and overly saturated results. 

To train a diffusion model on \DataName, we use the HuggingFace implementation of StableDiffusion v1.5 with $512\times 512$ resolution with the ControlNet adaptor. We keep the diffusion model frozen and finetune the ControlNet adaptor for $250,000$ iterations on a single A6000 GPU with batch size 4, starting learning rate $10^{-5}$, and a default DDPM noise scheduler. 

For layered score distillation sampling, we freeze both the diffusion model and the ControlNet adaptor, and only finetube the convolutional UNet generator producing the editing layers. We train it for $700$ steps with batch size $4$ for a single image, with a different timestep $t$ randomly sampled for each example in a batch, which achieves faster convergence. We observed that, contrary to the default score distillation sampling that requires a large classifier-free guidance (CFG) scale to achieve a well defined sampling, a GFC scale no more than 15 leads to visually more realistic relighting results. In our experiments, we use CFG $=10$ for the Dreambooth dataset, CFG$=7$ for the \DataName~ test set and CFG$=12$ for the Digital Art dataset. Additionally, we use an $L_1$ regularization term to minimize the over-exposure artifacts. For all the datasets, we use the following prompt to learn the editing layers: \textit{``A photo of a \{category name\} with \{X\} lighting"}, with the corresponding object category name and light direction index $X$. 

\subsection{Colorization}
  For colorization, we use a single-branch convolutional UNet generator with the same configuration as in the relighting experiments. The generator produces a single RGBA editing layer that determines the color overlay with the input sketch. 
  Since an off-the-shelf Stable Diffusion model already has a strong prior for colorization, we do not perform finetuning and use the default model instead. 

  To achieve a more visually pleasing and diverse translation results, we use a Variational Score Distillation (VSD)~\cite{wang2023prolificdreamer} that achieves superior text-to-3D generation results. VSD trains an additional LoRA adaptor to predict a denoising score for the 3D rendering result. Following this approach, we use a ControlNet adaptor instead of LoRA to predict a denoising score for the colorization result \emph{conditioned on the input sketch}. We train the generator for $4000$ iterations with the CFG scale of $8$, and we use an additional structure similarity loss based on the CLIP image encoder features as introduced in Clipasso~\cite{vinker2022clipasso} with the loss multiplyer of $2000$. 
\label{sec:app_colorization}

\subsection{Baselines}
\label{sec:app_baselines}

\xhdr{Text2Live}
We have two versions: one with a green screen (as default in the paper, and one without a green screen loss since shading as an image layer is hard to describe in text. For the latter, we omit the green screen loss in the total loss calculation. 
All configs are kept the same as recommended in the official implementation. 
For the text describing the green screen, we make it the object name. If we are omitting the green screen loss, we leave it blank. 
For the text describing the full edited image, we make a prompt: ``a digital art of a [object name] shaded with light coming from the left.''
For the text describing the input image: ``a digital art of a [object name]."

\xhdr{Prompt2Prompt}
We set the cross-replace weight to be 0.8, and the self-replace weight to be 0.6. All other parameters are as default in official implementation. 

\xhdr{DDS: Delta Denoising Score} 
We now present the setup for the other two baseline methods, for the DDS method: we use a guidance scale of 7.5, with minimum and maximum number of time steps to be $50$ and $1050$ respectively. The total number of iterations are 500 for each image optimization. The text edit prompt we use in this case are \textit{"a photo of [object] with dramatic light from left"} for \textit{ReLiT and DreamBooth} or \textit{"a digital art of [object] with dramatic light from left"} for \textit{Generated Art Datatset}. At first we keep the values as in the official implementation of the DDS. The the values of the time step and other configurations are obtained using a grid search approach.  

\xhdr{InstructPix2Pix}
Similar to DDS,  in the InstructPix2Pix we first keep the configuration as given in the official implementation and then change accordingly using the grid search approach. The guidance scale for text is 7.5 with the guidance scale on image as 1.5. The number of time steps used here are 120. \textit{``Add dramatic light from left on it"} is the text prompt used in this particular baseline for all three datasets as presented in the next section.

\section{Stable Diffusion lighting prior limitations}
\label{sec:app_prior_lim}

\section{Datasets}
\label{sec:app_datasets}

\subsection{ReLiT}
\label{sec:app_objaverser}
An illustration of the examples from the proposed ~\DataName~ dataset can be found in Figure~\ref{fig:objaverser}. \DataName~ contains $13975$ distinct training set objects and $164$ test set objects lit with the point light at $12$ predefined light source locations around the object from atop. In addition to the renderings, ~\DataName~ includes a metadata file with object category name and captions. The \DataName~ dataset will be released upon paper acceptance. 

\subsection{Generated Digital Art}
\label{sec:app_digital}
We generated the synthetic Digital Art dataset using the Stable Diffusion V2.1~\cite{rombach2022high} using the text prompt \textit{``A minimal digital art of a \{category\}"} for the following $13$ categories: birthday cake, car, cat, chair, corgi, cup, hamburger, high heel, ice cream, ice pop, orange, strawberry cheesecake, turtle. The Digital Art dataset will be released upon paper acceptance. 

\section{More results}
\label{sec:app_baselines}
\begin{figure*}
    \centering
    \includegraphics[width=1.1\linewidth]{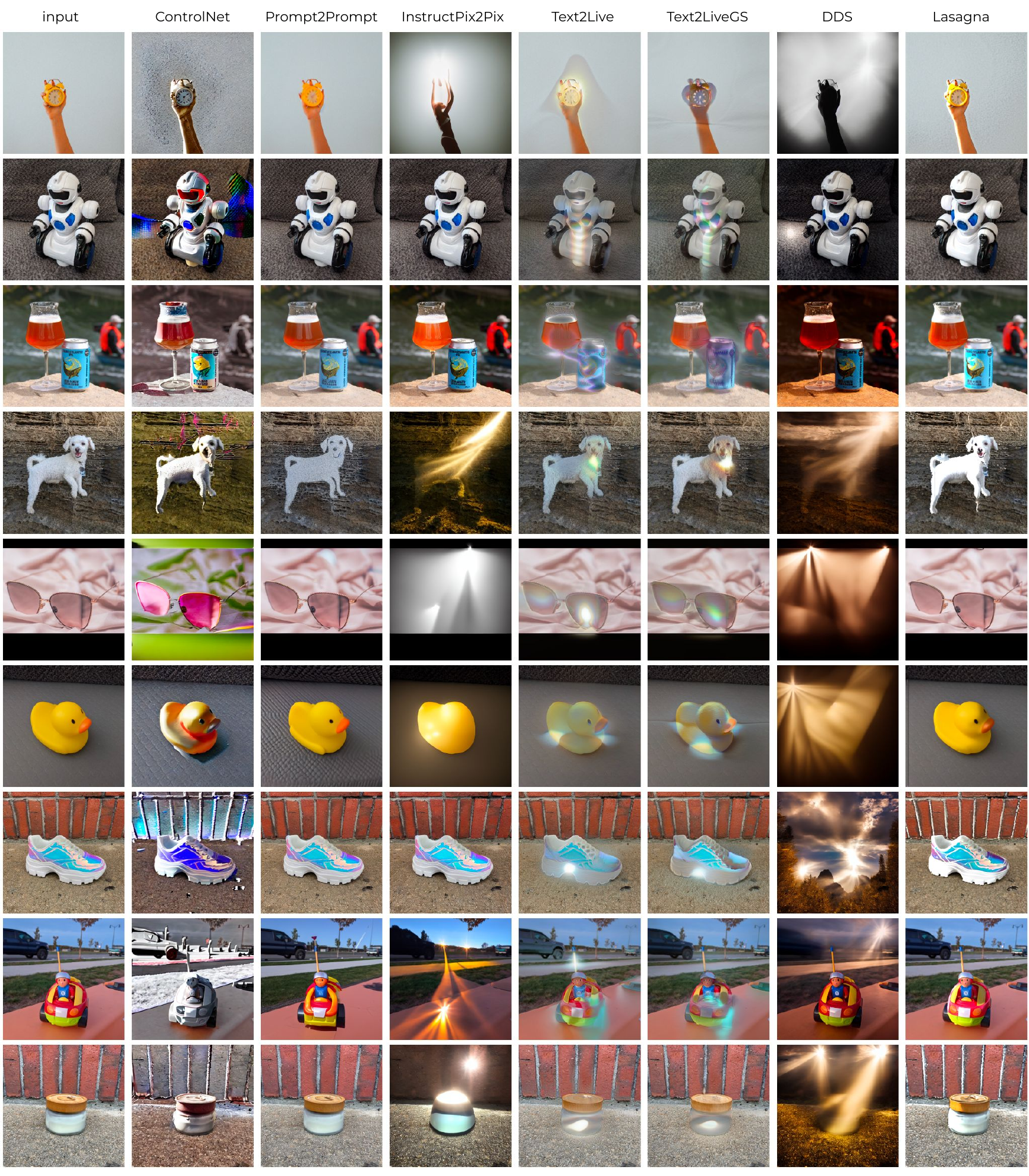}
    \caption{More \emph{randomly sampled} relighting examples on the DreamBooth data (instructed to relight from top left).}
    \label{fig:app_more_ex_db}
\end{figure*}

\begin{figure*}
    \centering
    \includegraphics[width=1.1\linewidth]{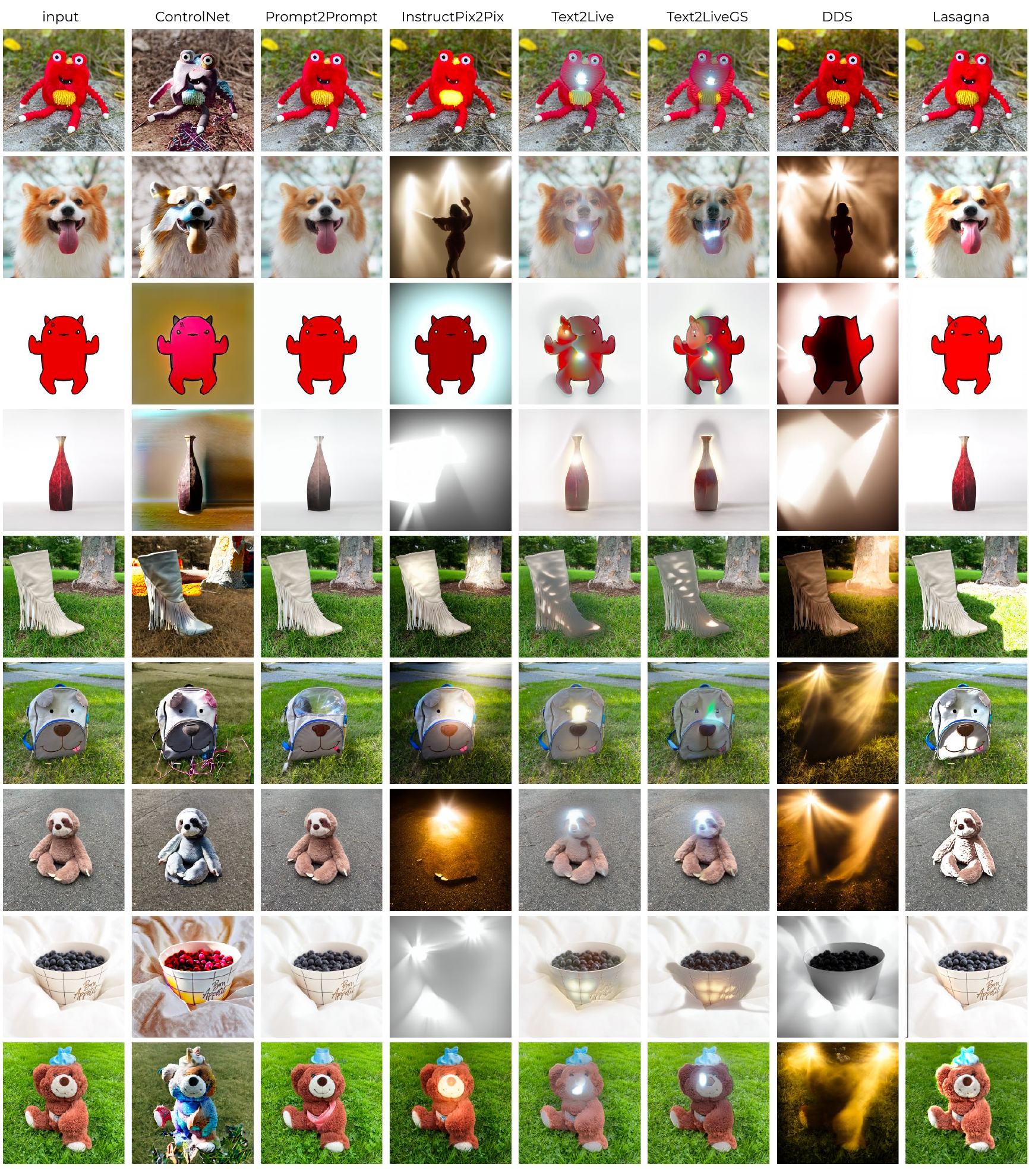}
    \caption{More \emph{randomly sampled} relighting examples on the DreamBooth data (instructed to relight from top left).}
    \label{fig:app_more_ex_db_2}
\end{figure*}

\begin{figure*}
    \centering
    \includegraphics[width=1.1\linewidth]{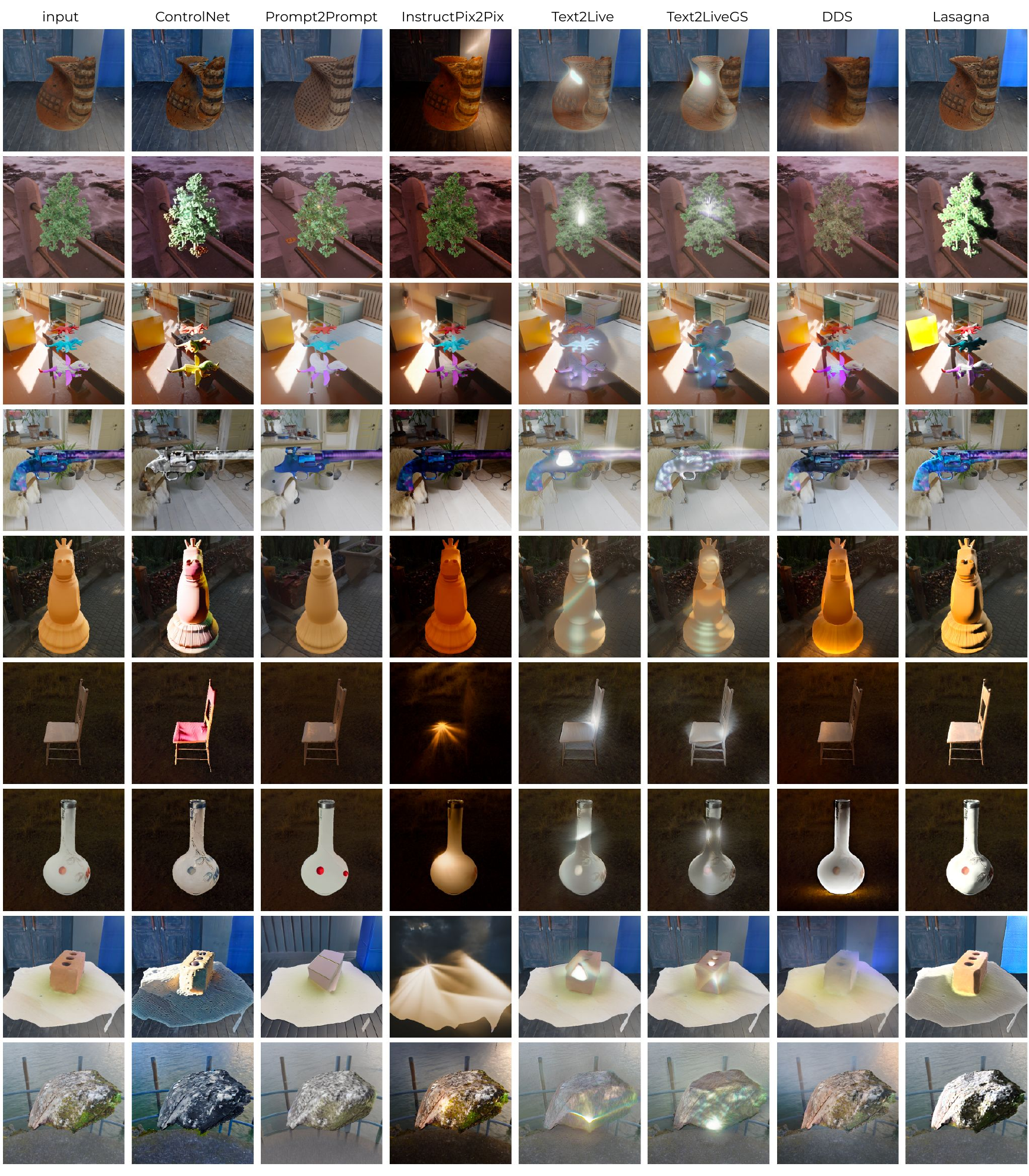}
    \caption{More \emph{randomly sampled} relighting examples on the \DataName~ test set data (instructed to relight from top left).}
    \label{fig:app_more_ex_relit_2}
\end{figure*}

\begin{figure*}
    \centering
    \includegraphics[width=1.1\linewidth]{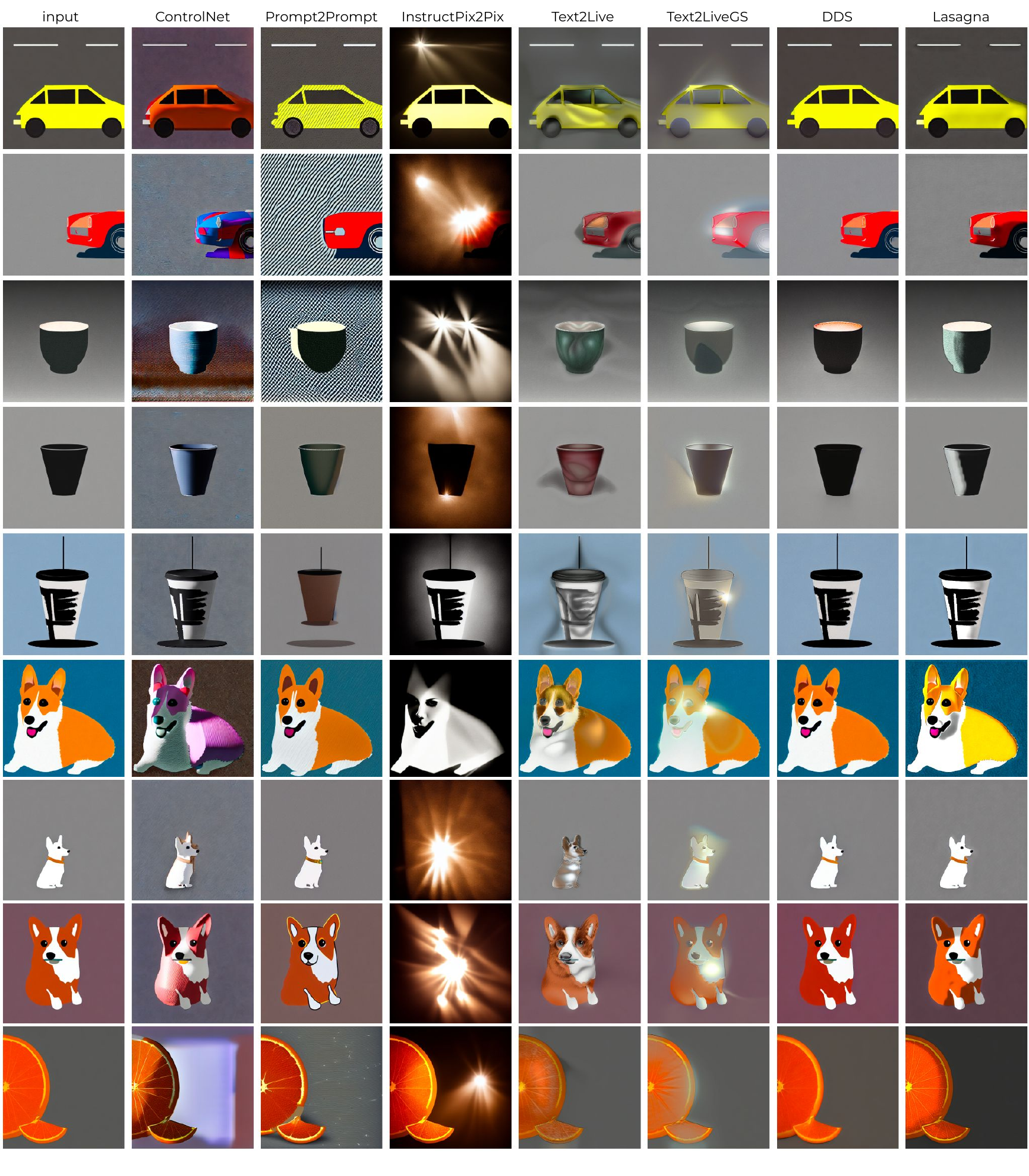}
    \caption{More \emph{randomly sampled} relighting examples on the Digital Art data (instructed to relight from top left).}
    \label{fig:app_more_ex_digital}
\end{figure*}

\begin{figure*}
    \centering
    \includegraphics[width=1.1\linewidth]{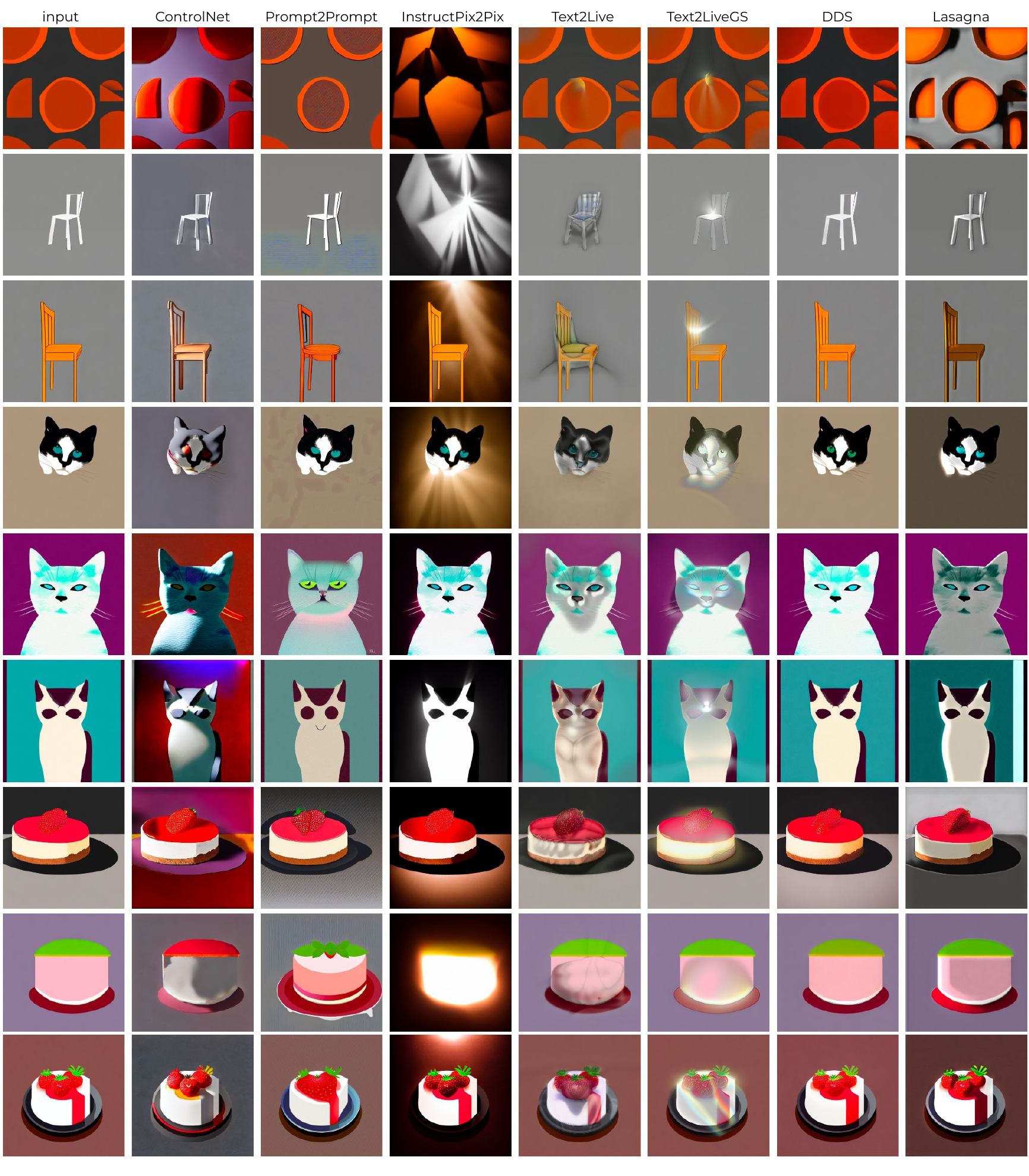}
    \caption{More \emph{randomly sampled} relighting examples on the Digital Art data (instructed to relight from top left).}
    \label{fig:app_more_ex_digital_2}
\end{figure*}

\begin{figure*}
    \centering
    \includegraphics[width=1.1\linewidth]{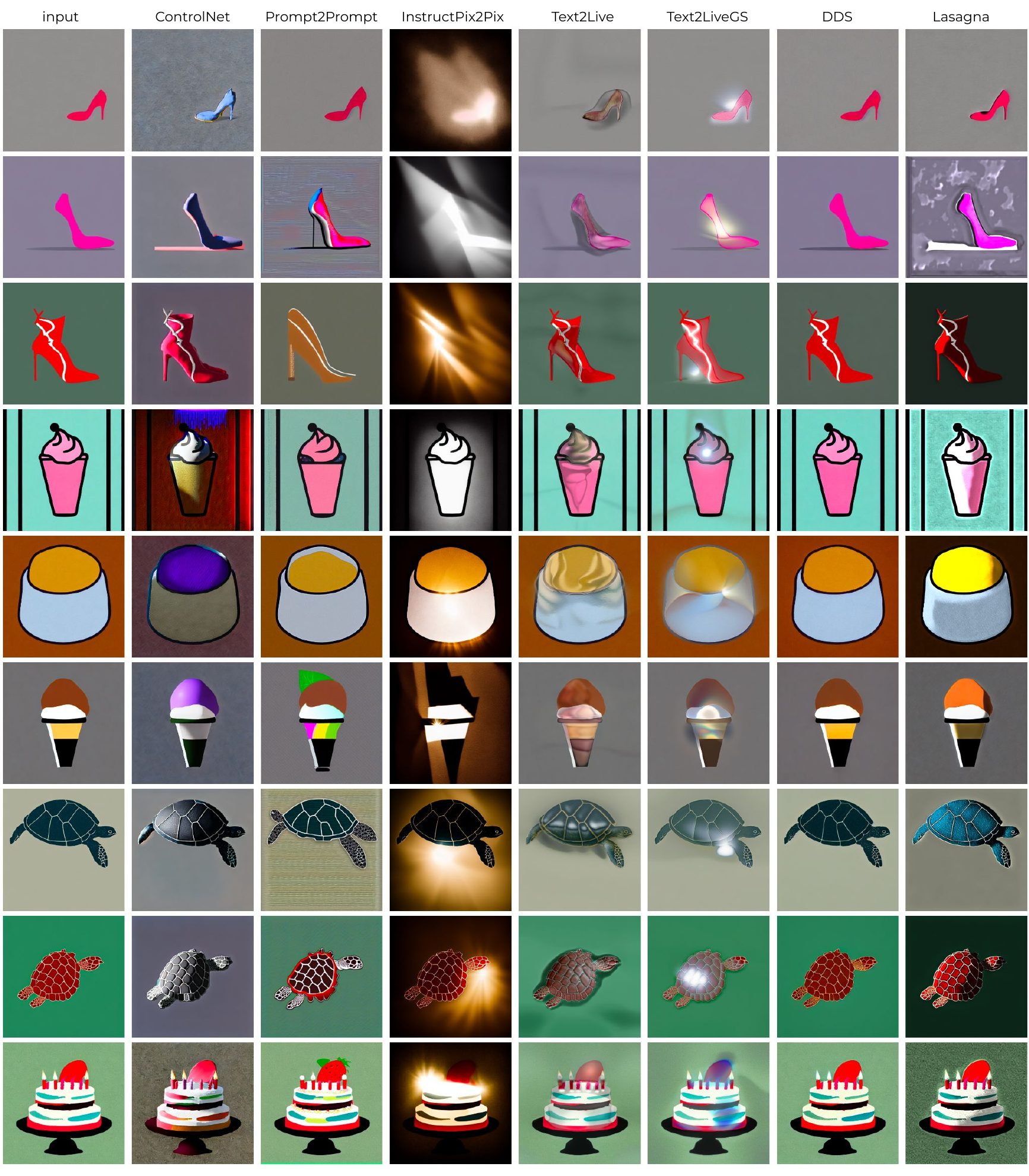}
    \caption{More \emph{randomly sampled} relighting examples on the Digital Art data (instructed to relight from top left).}
    \label{fig:app_more_ex_digital_3}
\end{figure*}

\begin{figure*}
    \centering
    \includegraphics[width=1.1\linewidth]{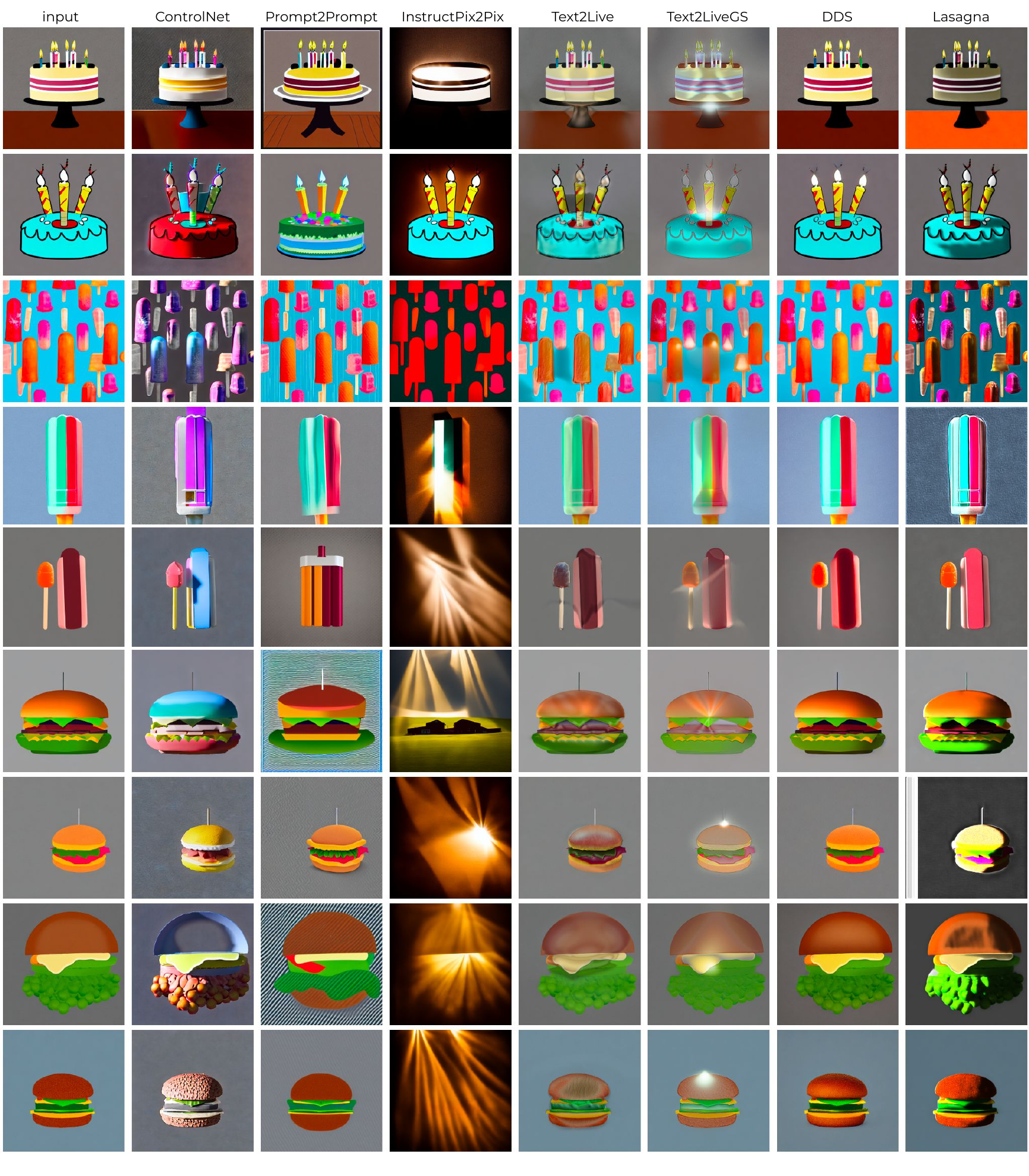}
    \caption{More \emph{randomly sampled} relighting examples on the Digital Art data (instructed to relight from top left).}
    \label{fig:app_more_ex_digital_4}
\end{figure*}

More relighting examples by \MethodName~ and the baseline methods can be found in Figures~\ref{fig:app_more_ex_db} and ~\ref{fig:app_more_ex_db_2} for the DreamBooth dataset, in Figures~\ref{fig:app_more_ex_relit} and ~\ref{fig:app_more_ex_relit_2} for the test set of \DataName~ dataset, and on Figures~\ref{fig:app_more_ex_digital},~\ref{fig:app_more_ex_digital_2},~\ref{fig:app_more_ex_digital_3} and ~\ref{fig:app_more_ex_digital_4} for the Digital Art dataset.
\begin{figure*}
    \centering
    \includegraphics[width=1.1\linewidth]{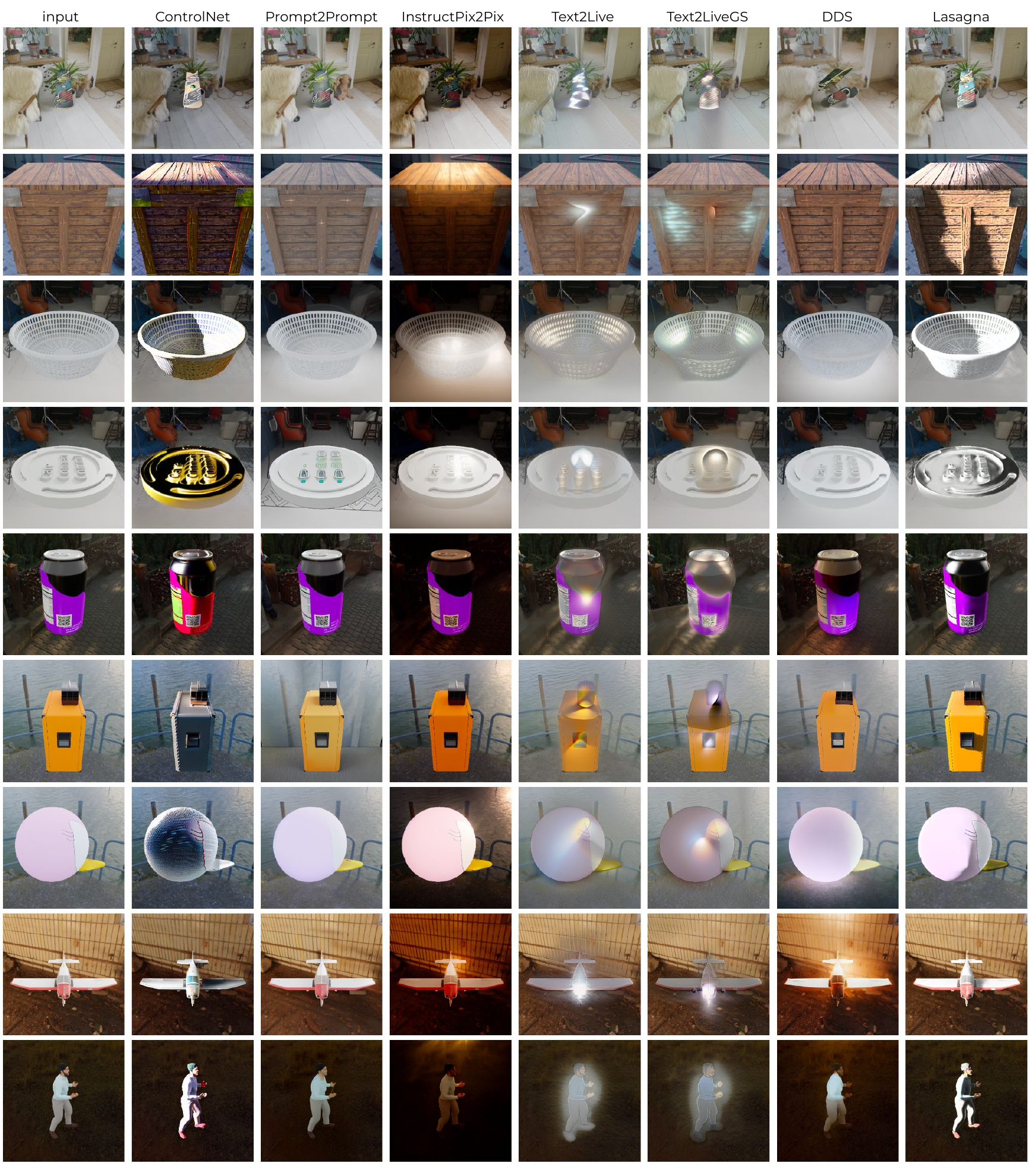}
    \caption{More \emph{randomly sampled} relighting examples on the \DataName~ test set data (instructed to relight from top left).}
    \label{fig:app_more_ex_relit}
\end{figure*}

\begin{figure*}
    \centering
    \includegraphics[width=1.1\linewidth]{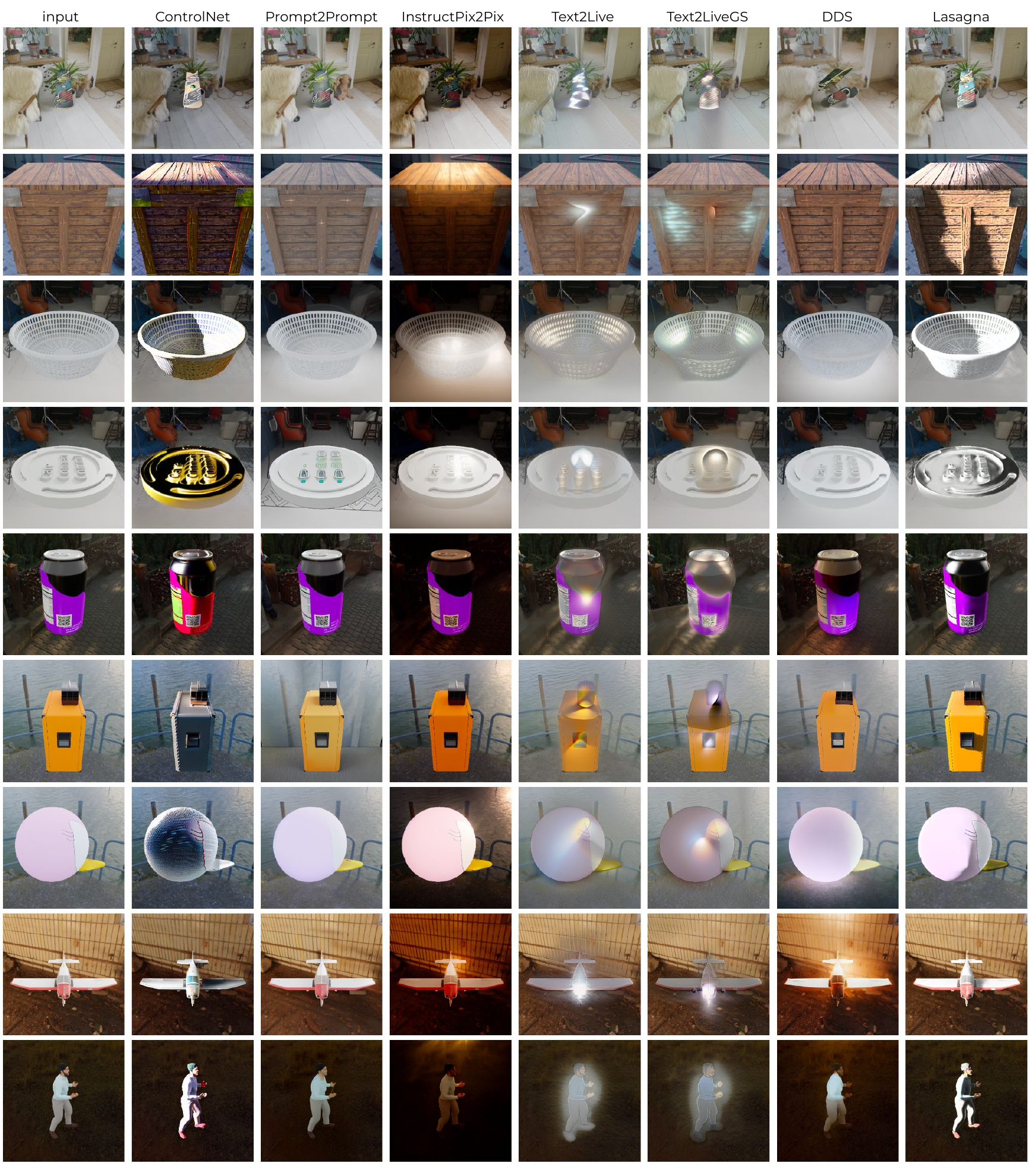}
    \caption{More \emph{randomly sampled} relighting examples on the \DataName~ test set data (instructed to relight from top left).}
    \label{fig:app_more_ex_relit}
\end{figure*}

\section{Human Evaluation Details and Illustrations}
\label{sec:app_human_eval}
\begin{figure*}
    \centering
    \includegraphics[width=\linewidth]{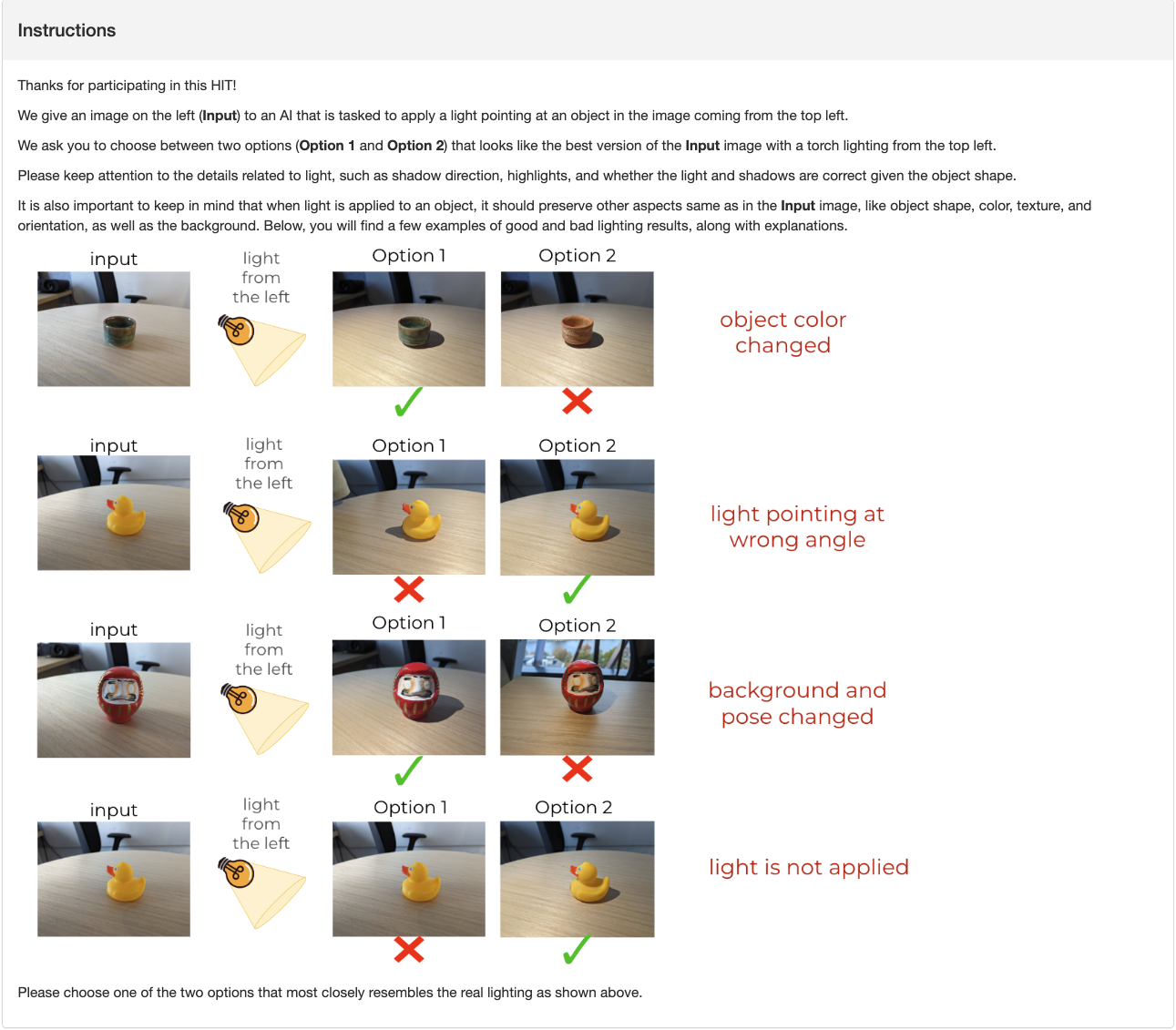}
    \caption{Instructions for the human evaluation task. }
    \label{fig:app_human_eval_instructions}
\end{figure*}

\begin{figure*}
    \centering
    \includegraphics[width=0.9\linewidth]{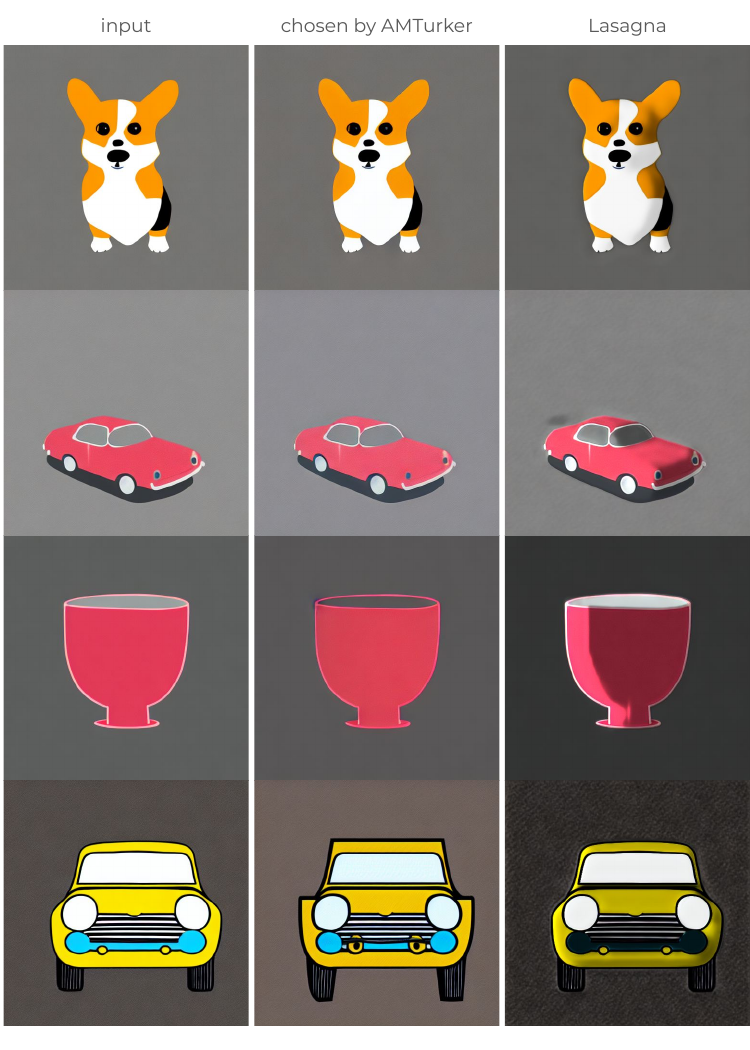}
    \caption{Illustration of the relighting choice made by the participants in a human evaluation study performed on the AMT indicates how hard the evaluating for relighting is for a participants without a specialized training and expertise.}
    \label{fig:app_amt_eval}
\end{figure*}

\begin{figure*}
    \centering
    \includegraphics[width=0.9\linewidth]{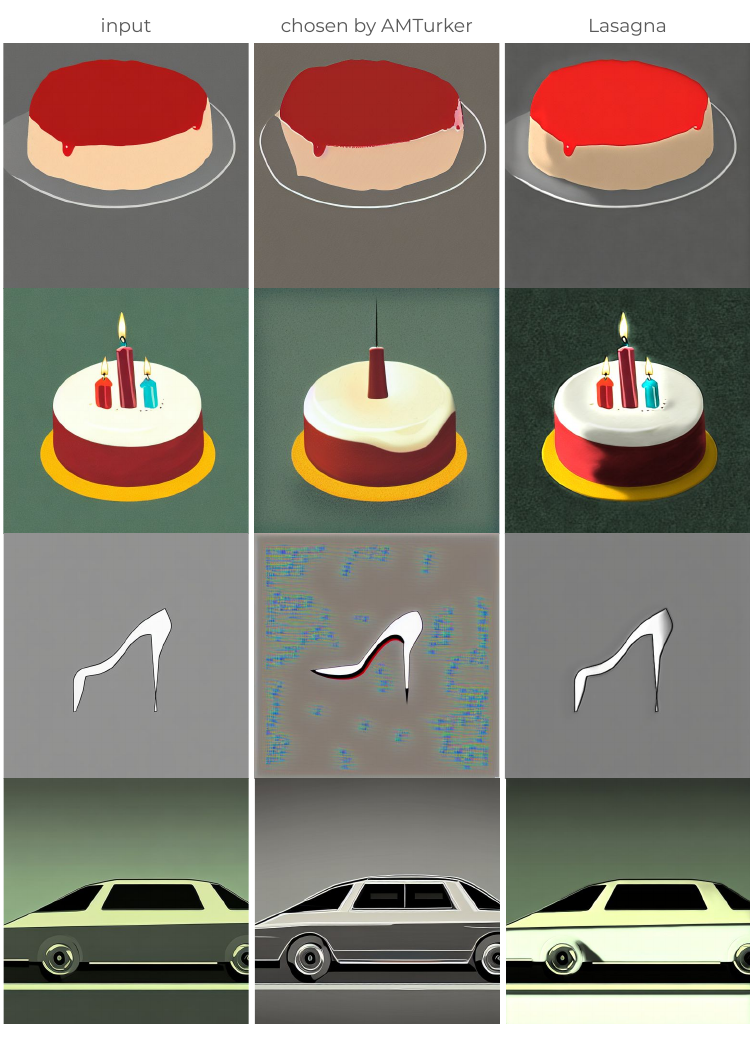}
    \caption{Illustration of the relighting choice made by the participants in a human evaluation study performed on the AMT indicates how hard the evaluating for relighting is for a participants without a specialized training and expertise.}
    \label{fig:app_amt_eval_2}
\end{figure*}

\begin{figure*}
    \centering
    \includegraphics[width=0.9\linewidth]{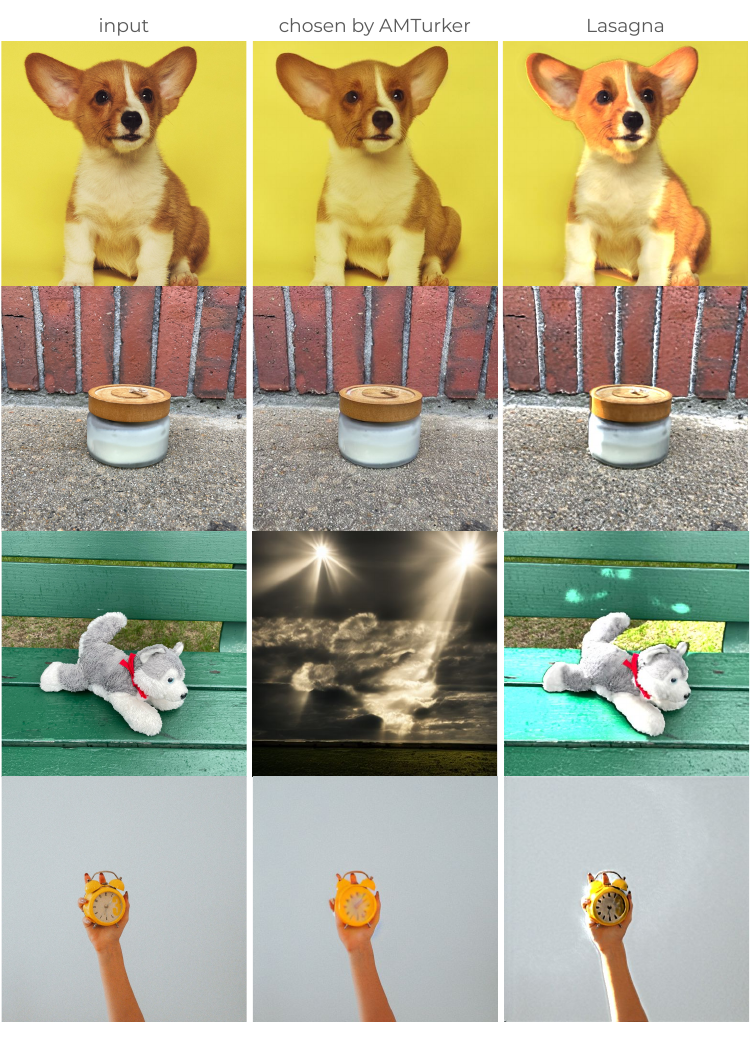}
    \caption{Illustration of the relighting choice made by the participants in a human evaluation study performed on the AMT indicates how hard the evaluating for relighting is for a participants without a specialized training and expertise.}
    \label{fig:app_amt_eval_3}
\end{figure*}

\begin{figure*}
    \centering

    % Subfigure 1
    \begin{subfigure}{0.33\textwidth}
    \centering
        \includegraphics[width=\textwidth]{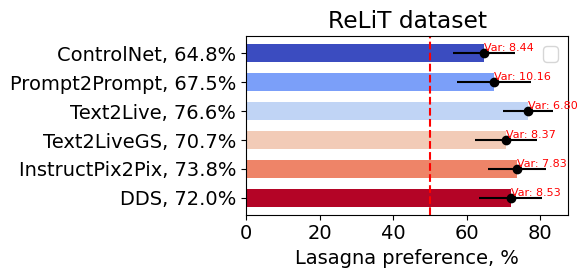}
        \caption{\DataName~ test set}
    \end{subfigure}
    % Subfigure 2
    \begin{subfigure}{0.33\textwidth}
        \centering
        \includegraphics[width=\textwidth]{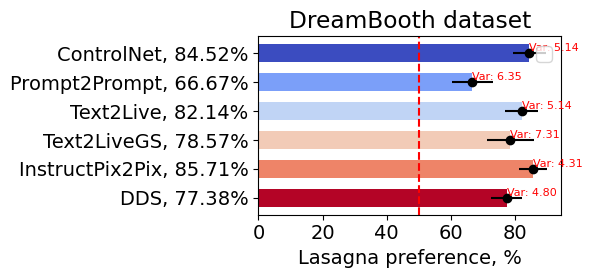}
        \caption{DreamBooth dataset}
    \end{subfigure}
    % Subfigure 3
    \begin{subfigure}{0.33\textwidth}
        \centering
        \includegraphics[width=\textwidth]{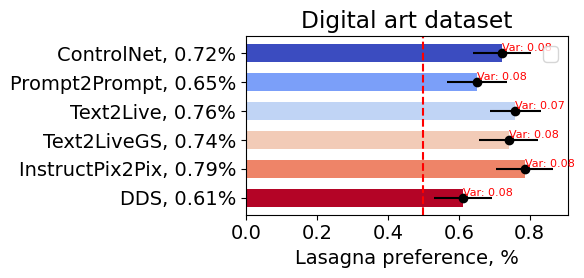}
        \caption{Digital Art}
    \end{subfigure}

    \caption{ Percentage of cases in which the participants chose \MethodName~ over each baseline considered in the human study conducted via AMTurk. }
    \label{fig:app_amt_study}
\end{figure*}

Instructions given to the human evaluation participants can be found in Figure~\ref{fig:app_human_eval_instructions}. 
Our human evaluation results on Amazon Mechanical Turk illustrated in Figure~\ref{fig:app_amt_study} indicate that relighting is an edit type that is hard to evaluate without expertise and special training. As illustrated in Figures~\ref{fig:app_amt_eval}, ~\ref{fig:app_amt_eval_2} and ~\ref{fig:app_amt_eval_3}, participants of the AMTurk study that were given the instructions in Figure~\ref{fig:app_human_eval_instructions}, often favor the results that are almost identical to the input image and do not present any changes in the lighting to the \MethodName~ results with a valid lighting. We hypothesise that the participants are making such choices since the direct replica of the input image is often perceived as a more realistic looking image overall.

In order to mitigate such a bias towards the overall perceived realism on the relighting quality evaluation, we conduct an expert human evaluation study including three participants with the computer science background. The participants had no prior knowledge of how each of the evaluated methods editing results look like, and were given the same instructions as the participants of the AMTurk study (Figure~\ref{fig:app_human_eval_instructions}). The results of the expert evaluation study have a significantly higher expert agreement range across the examples, with the variance of the Two-Forced-Choice scores being over $3.5$ times lower than that of the AMTurk evaluation study, suggesting a more reliable assessment. 

\paragraph{Automatic metrics} such as mean square error (MSE) or LPIPS score w.r.t. the ground truth relit images from the test set of \DataName~, as shown in Table~\ref{tab:objaverse}, tend to favor the results that are perceptually indistinguishable from the input images, such as Prompt2Prompt or DDS translation (please see the results illustrated in Figures~\ref{fig:app_more_ex_db}-~\ref{fig:app_more_ex_digital_4}). Since these metrics are more sensitive to the general changes in w.r.t. the input image and do not accurately reflect the translation accuracy, we opt for a human evaluation instead.

\begin{table*}\small
\begin{tabular}{c c c}
\toprule
     & MSE $\downarrow$ &  LPIPS$\downarrow$  \\ \midrule
     Prompt2Prompt &  $\bm{0.0123 \pm 0.0001}$ & $6.33 \times 10^{-5} \pm 2\times 10^{-9}$ \\
     InstructPix2Pix &   $0.05 \pm 0.0001$ & $9.97\times 10^{-5}\pm 6\times 10^{-9}$\\
     Text2Live &  $0.0166 \pm 0.0001$ &  $7.4\times 10^{-5} \pm 3\times 10^{-9}$\\
     Text2Live $\lambda_{GS}  = 0$ &  $0.0164 \pm 0.0002$ & $\bm{6.8\times 10^{-5}\pm 2\times 10^{-9}}$ \\
     DDS & $0.0156 \pm{0.0001}$  & $6.9\times 10^{-5}\pm 2\times 10^{-9}$ 
     \\ \midrule
    ControlNet &   $0.0349 \pm 0.002$ & $1.5 \times 10^{-4} \pm 2\times 10^{-9}$\\
    \MethodName  & $0.0159 \pm 0.001$ & $7.2\times 10^{-5} \pm 8\times 10^{-9}$\\
    
    \bottomrule
\end{tabular}
    \caption{Relighting accuracy (mean squared error) on the subset of \DataName~ training and test sets with ControlNet and \MethodName. }
    \label{tab:objaverse}
\end{table*}


\end{document}